\documentclass[a4paper]{llncs}

\usepackage{blindtext}
\usepackage{url}            
\usepackage{booktabs}       
\usepackage{amsfonts}       
\usepackage{nicefrac}       
\usepackage{microtype}      
\usepackage{xcolor}         
\usepackage{natbib}
\usepackage{graphicx}
\usepackage{lipsum}  
\usepackage{xspace}
\usepackage[center]{subfigure}
\usepackage{amsmath}
\usepackage{hyperref}
\usepackage{tikz,ifthen}
\usepackage{graphicx}

\definecolor{lightgrey}{rgb}{.99,.99,.99}
\definecolor{myred}{rgb}{.76,.07,.12}
\definecolor{lapisblue}{rgb}{.15,.38,.61}

\usetikzlibrary{arrows,trees,backgrounds,automata,shapes,decorations,plotmarks,fit,calc,positioning,shadows,chains}
\definecolor{color0}{RGB}{239, 48, 84} 
\definecolor{color4}{RGB}{62, 180, 139} 
\definecolor{color6}{RGB}{147, 129, 255} 
\definecolor{color7}{RGB}{242, 193, 78} 

\definecolor{nnedgecolor}{RGB}{90,90,90}
\tikzstyle{every pin edge}=[<-,shorten <=1pt]
\tikzstyle{every path}=[draw=color7!50]
\tikzstyle{neuron}=[circle,fill=black!25,minimum size=17pt,inner sep=0pt]
\tikzstyle{input neuron}=[neuron, fill=color4]
\tikzstyle{output neuron}=[neuron, fill=red!50]
\tikzstyle{hidden neuron}=[neuron, fill=blue!50]
\tikzstyle{annot} = [text width=4em, text centered]
\tikzstyle{nnedge} = [-{stealth},shorten >=0.1cm, shorten <=0.05cm,line 
width=0.8pt,nnedgecolor]
\usetikzlibrary{calc}

\newcommand{\sat}{\texttt{SAT}}
\newcommand{\unsat}{\texttt{UNSAT}}
\newcommand{\turtleenv}{\texttt{Robotic Mapless Navigation}}
\newcommand{\turtleenvshort}{\texttt{Mapless Navigation}}
\newcommand{\gridenv}{\texttt{Jumping World}}

\newcommand{\successrate}{\texttt{Success Rate}}
\newcommand{\collisionrate}{\texttt{Collision Rate}}
\newcommand{\fvcollisionrate}{\texttt{Adv.~Collision Rate}}
\newcommand{\adversarialrate}{\texttt{Adversarial Rate}}

\newcommand{\prove}{\texttt{ProVe}}
\newcommand{\countingprove}{\texttt{Counting-ProVe}}

\newcommand{\actionSpace}{\mathcal{A}}
\newcommand{\statesSpace}{\mathcal{S}}

\begin{document}


\title{Analyzing Adversarial Inputs in\\ Deep Reinforcement Learning}

    
	\author{
        Davide Corsi\inst{1} \and
		Guy Amir\inst{2} \and
		Guy Katz\inst{3} \and
        Alessandro Farinelli\inst{4}
	}
	\institute{
         University of California, Irvine, USA\\
           \email{dcorsi@uci.edu}
          \\
            \and
          Cornell University, USA\\
           \email{gda42@cornell.edu}
          \\
          \and
        The Hebrew University of Jerusalem, Israel\\ \email{guykatz@cs.huji.ac.il} \and University of Verona, Italy\\
           \email{alessandro.farinelli@univr.it}
	}
	
\maketitle

\begin{abstract}
In recent years, Deep Reinforcement Learning (DRL) has become a popular paradigm in machine learning due to its successful applications to real-world and complex systems. However, even the state-of-the-art DRL models have been shown to suffer from reliability concerns --- for example, their susceptibility to adversarial inputs, i.e., small and abundant input perturbations that can fool the models into making unpredictable and potentially dangerous decisions. This drawback limits the deployment of DRL systems in safety-critical contexts, where even a small error cannot be tolerated. In this work, we present a comprehensive analysis of the \emph{characterization} of adversarial inputs, through the lens of formal verification. Specifically, \textcolor{black}{we present the \adversarialrate{}, a metric adapted from the ProVe family~\citep{ProVe}, for the systematic evaluation of adversarial inputs in DRL, which partitions the input domain into subregions to enable both quantification and spatial visualization of adversarial inputs. The main contribution of this work is to provide a comprehensive evaluation framework for the effect of adversarial inputs on DRL policies.} We present a set of tools and algorithms for its computation. Our analysis empirically demonstrates how adversarial inputs can affect the safety of a given DRL system with respect to such perturbations. Moreover, we analyze the behavior of these configurations to suggest several useful practices and guidelines to help mitigate the vulnerability of trained DRL networks. 
\end{abstract}


\section{Introduction}

In the past decade, Deep Reinforcement Learning (DRL) has achieved remarkable success in a wide range of tasks, including robotics \citep{akkaya2019solving}, protein folding \citep{jumper2021highly}, game playing \citep{mnih2013playing}, and more \citep{wurman2022outracing, clarke2020deep}. However, Deep Learning models are known to suffer from susceptibility to \emph{adversarial attacks} --- small perturbations to their inputs, which may cause them to err and behave unexpectedly \citep{szegedy2013intriguing}. This vulnerability is observed even in state-of-the-art DRL agents \citep{amir2023verifying}, and limits their deployment in safety-critical domains such as robotic control \citep{ibarz2021train}, autonomous driving \citep{bojarski2016end}, and healthcare \citep{pore2021safe, corsi2023constrained}, where even a single error may have dire consequences. 
To address this critical issue, many tools and algorithms have been developed to improve the reliability of DRL agents. Some of these techniques are incorporated during the DRL training phase to robustify the trained DRL agents, including constrained reinforcement learning \citep{achiam2017constrained, CoYeAm22}, safe exploration \citep{YaSiJa22, SrEyHa20}, and various reward-shaping techniques \citep{ChDaSh19}. Other methods are based on formal verification \citep{KaBaDiJuKo17, wang2021beta}, i.e., automatic and rigorous techniques that are applied after the training phase to automatically validate whether a given DRL agent is formally safe to deploy.

Although these approaches present promising results, they have many setbacks. Specifically, the popular training-based methods typically provide safer behavior only in expectation, without directly analyzing the underlying problem of the model's susceptibility. Formal methods, on the other hand, do provide some level of rigor, but typically do not characterize the \textit{global characteristics} of adversarial perturbations that they identify. Hence, there is a need for methods that not only provide formal guarantees regarding DRL robustness but also \emph{characterize} the adversarial inputs in a given state space. For example, understanding not only if such adversarial inputs exist, but how they are dispersed in the input space of a DRL agent and how they evolve in relation to the training techniques, network structures, and environments. Such an understanding is key for generating more reliable DRL systems, and hence allowing their deployment in various safety-critical domains. An additional gap in the literature is the lack of methods for assessing adversarial inputs within the specific context of DRL agents, as most methods are geared towards general Machine Learning models \citep{henderson2018deep, tian2020evaldnn}.

In this work, we make a step towards bridging this gap, by presenting a comprehensive analysis regarding the impact of adversarial inputs on DRL agents. Building upon our previous work in the domain of DNN verification \citep{amir2023verifying, corsi2023constrained, pore2021safe, CountingProVe} and particularly on our tool \texttt{ProVe} \citep{ProVe} (which is based on the backend verification technology of \texttt{Reluplex} \citep{KaBaDiJuKo17} and \texttt{Marabou} \citep{katz2019marabou, WuIsZeTaDaKoReAmJuBaHuLaWuZhKoKaBa24}), \textcolor{black}{we present the \adversarialrate{}, a metric building upon the \texttt{Violation Rate} introduced in the prior work from Corsi et al.~\citep{ProVe} and adapted for the systematic evaluation of adversarial inputs in the DRL domain. Specifically, while the original \texttt{Violation Rate} measures the overall size of the unsafe portion of the input space, the \adversarialrate{} is designed to partition the input domain into subregions, enabling fine-grained spatial visualization and analysis of adversarial inputs alongside the environment layout. The underlying verification procedure is further extended to enumerate violating input points and expose them for downstream empirical evaluation via the \fvcollisionrate{}. The main contribution of this work is therefore not to introduce a new verification tool, but to provide a systematic and comprehensive evaluation framework for studying the effect of adversarial inputs on reinforcement learning policies.} In this work, we extensively demonstrate the usefulness of the \adversarialrate{} in assessing the robustness of various DRL agents, in multiple scenarios, as well as put forth a set of tools for its computation. We also show how the \adversarialrate{} can identify adversarial inputs that could otherwise be undetected with conventional methods. 
The core of this work includes an in-depth analysis of the effect of adversarial inputs on the behavior of DRL systems. Such analysis aims at answering the following, important research questions: \textit{Are adversarial inputs a concrete threat to state-of-the-art DRL systems?} \textit{Do adversarial inputs depend on training procedures?} \textit{Is it possible to limit the pervasiveness of adversarial inputs by user-specific design choices?} Specifically, we consider several aspects related to the impact of adversarial inputs, including:

\begin{itemize}
    \item A \emph{spatial} analysis of adversarial inputs --- in which we demonstrate the existence of \emph{unsafe regions}, i.e., subdomains of the input space, in which the DRL agent is more likely to exhibit faulty behavior than in others. We also analyze the relations and similarities between unsafe regions among \emph{different} models trained on the same task.

    \item A \emph{temporal} analysis --- in which we analyze the overall dependency of adversarial inputs throughout  the training process. Specifically, we show how the unsafe regions can ``move'' throughout the training iterations within unpredictable patterns.

    \item A \emph{model-specific} analysis --- in which we analyze how the architecture of the DRL agent can affect the agent's susceptibility to adversarial inputs. More precisely, we inquire about the correlation between adversarial inputs and the agent's \emph{size} (i.e., the number of neurons in the neural network) and \emph{expressibility} (i.e., the type of activation functions).
\end{itemize}

In our evaluation, we extensively train thousands of agents on two popular DRL tasks:  \gridenv{} (a variant of the GridWorld benchmark) and \turtleenvshort{}. \gridenv{} is a classic benchmark in which an agent moves on a two-dimensional grid towards a goal \citep{MinigridMiniworld23, chevalier2018babyai}; we rely on \gridenv{} as a running example to clarify the main concepts related to our analysis, due to its simplicity and intuitive setting. The \turtleenvshort{} benchmark complements our analysis by demonstrating our results on the complex task of robotic navigation in a real-world setting \citep{chiang2019learning}. We trained the agents with state-of-the-art algorithms, namely PPO \citep{ShWoDh17} and TD3 \citep{fujimoto2018addressing}, as they are widely considered two of the most popular and efficient DRL algorithms.

Our analysis presents surprising results. For example, we show that adversarial inputs may be particularly concentrated in certain small regions of the input space, and thus unlikely to be discovered in a standard empirical evaluation phase (while efficiently discovered by our \adversarialrate{} metric). Nevertheless, we show how few adversarial inputs have a major impact on the safety of the system. We also present additional findings, on which we extensively elaborate in other sections.
To the best of our knowledge, this is the first work that comprehensively analyzes the properties of adversarial inputs in DRL through the lens of formal methods. We hope, and believe, that this work will pave the way for future research to come and address the timely problem of DRL safety, and specifically, DRL robustness against adversarial inputs. 

The rest of this paper is organized as follows. Sec.~\ref{sec:background} contains background and related work on DNNs, DRLs, and their verification procedure. In Sec.~\ref{sec:tools} we present the various formal methods and tools, used in this work. We elaborate on our benchmarks in Sec.~\ref{sec:benchmarks}, and present our extensive analysis in Sec.~\ref{sec:analysis}. In Sec.~\ref{sec:retraining} we discuss some existing techniques to robustify DRL agents. Finally, we conclude our results and discuss potential future directions in Sec.~\ref{sec:conclusion}.
\section{Background and Related Work} 
\label{sec:background}

In this section, we introduce the fundamental concepts necessary for the reader. In particular, we present the definitions of Deep Neural Network (DNNs) and Deep Reinforcement Learning (DRL). Subsequently, we define the formal verification problem of DNNs, and its extension to the \#DNN-verification problem, which constitutes the basis for this work.

\subsection{Deep Neural Networks (DNN)}
\begin{figure}[h]
    \begin{center}
    \def\layersep{2.0cm}
    \begin{tikzpicture}[shorten >=1pt,->,draw=black!50, node
    distance=\layersep,font=\footnotesize]
    
    \node[input neuron] (I-1) at (0,-1) {$v^1_1$};
    \node[input neuron] (I-2) at (0,-2.5) {$v^2_1$};
    
    \node[left=-0.05cm of I-1] (b1) {$[2]$};
    \node[left=-0.05cm of I-2] (b2) {$[-1]$};
    \node[hidden neuron] (H-1) at (\layersep,-1) {$v^1_2$};
    \node[hidden neuron] (H-2) at (\layersep,-2.5) {$v^2_2$};	
    \node[hidden neuron] (H-3) at (2*\layersep,-1) {$v^1_3$};
    \node[hidden neuron] (H-4) at (2*\layersep,-2.5) {$v^2_3$};
    \node[output neuron] at (3*\layersep, -1.75) (O-1) {$v^1_4$};
    
    \draw[nnedge] (I-1) --node[above] {$5$} (H-1);
    \draw[nnedge] (I-1) --node[above, pos=0.3] {$-1$} (H-2);
    \draw[nnedge] (I-2) --node[below, pos=0.3] {$-1$} (H-1);
    \draw[nnedge] (I-2) --node[below] {$3$} (H-2);
    \draw[nnedge] (H-1) --node[above] {ReLU} (H-3);
    \draw[nnedge] (H-2) --node[below] {ReLU} (H-4);
    \draw[nnedge] (H-3) --node[above] {$1$} (O-1);
    \draw[nnedge] (H-4) --node[below] {$3$} (O-1);
    
    \node[below=0.05cm of H-1] (b1) {$[11]$};
    \node[below=0.05cm of H-2] (b2) {$[-5]$};
    
    \node[below=0.05cm of H-3] (b1) {$[11]$};
    \node[below=0.05cm of H-4] (b2) {$[0]$};
    \node[right=0.05cm of O-1] (b1) {$[11]$};
    
    \node[annot,above of=H-1, node distance=0.8cm] (hl1) {Weighted sum};
    \node[annot,above of=H-3, node distance=0.8cm] (hl2) {Activation Layer };
    \node[annot,left of=hl1] {Input };
    \node[annot,right of=hl2] {Output };
    \end{tikzpicture}
    \caption{A toy DNN.}
    \label{fig:toy-net}
    \end{center}
\end{figure}
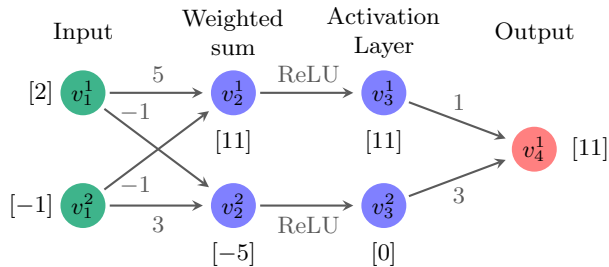

A DNN is a function $\mathcal{N}_\theta(x)$, defined over the parameters $\theta$, that maps an input vector $x$ to an output $y$. The computation is performed through a sequence of operations for each layer of the DNN:

\begin{equation}
    a^{(l)} = g(W^{(l)}a^{(l-1)}+b^{(l)})
\end{equation}

\noindent where: (i) $a^{(l)-1}$ is the input vector to layer $l$; (ii) the weights matrix $W$ and the biases $b$ are the parameters of the function (conventionally denoted together as $\theta = [W^{(0)}, ... W^{(n)}, b^{(0)}, ... b^{(n)}]$); and (iii) $g$ is a non-linear activation function. Notice that $a^{(0)}$ represents the input vector, and $a^{(n)}$ is the output. A toy example of a DNN is presented in Fig. \ref{fig:toy-net}.

In the common feed-forward architecture, each input is connected to subsequent layers, until reaching the (final) output layer. For example, if the toy DNN depicted in Fig. \ref{fig:toy-net}, is fed the input vector $[2, -1]^T$, the second layer computes the values $[11,-5]^T$. Then, the ReLU activations are applied, resulting in $[11,0]^T$. Finally, the last layer computes the single output value $[11]$. We also note that the \textit{activation functions} are a fundamental component of DNNs, and determine their expressivity, and also (as we elaborate in Sec.~\ref{sec:analysis}) the relation between this expressivity and their susceptibility to adversarial configurations. Many popular activation functions are piecewise-linear, e.g., \textit{ReLU} ($ReLU(x)=\max(0,x)$) and its variants (e.g.,  \textit{Leaky-ReLU}), other popular activations include \textit{sigmoid}, \textit{hyperbolic tangent}, and \textit{swish}. 

\subsection{Deep Reinforcement Learning (DRL)}
DRL stands as a prominent paradigm in machine learning, wherein a DNN-based reinforcement learning (RL) agent engages with an \textit{environment} over multiple time-steps \(t \in \{0,1,2,...\}\), with the aim of learning to map an input (from the state-space $\statesSpace$) to an appropriate output action (from the action-space $\actionSpace$). At each discrete time-step, the DRL agent observes the \textit{state} (\(s_{t} \in \statesSpace\)) of the environment, and chooses an \textit{action} (\(N(s_t)=a_{t} \in \actionSpace\)), which leads to the next state (\(s_{t+1}\)). Throughout the training, the environment provides a \textit{reward} (\(r_t\)) at each time-step, based on the previously chosen action. The agent undergoes training through repeatedly interacting with the environment, aiming to maximize its \textit{expected cumulative discounted reward} (\(R_t=\mathbb{E}\big[\sum_{t}\gamma^{t}\cdot r_t\big]\)), where \(\gamma \in [0,1]\) represents a \textit{discount factor} --- a hyperparameter regulating the cumulative impact of past decisions on the reward. For more details, we refer the reader to~\cite{sutton2018reinforcement}.

\subsection{Formal Verification of DNNs}
In recent years, a plethora of tools has been put forth to \emph{formally} and \emph{rigorously} verify the behavior of DNNs, which have an opaque, ``black-box'' nature. These methods attempt to verify that the DNN in question adheres to user-defined specifications for \emph{all possible inputs} in a given domain. Next, we present the formal definition of the DNN verification problem.

\begin{definition}[\textit{The DNN-Verification Problem}]
\label{def:decision_problem}
\phantom{a}

    {\bf Input}: A tuple $\mathcal{R}=\langle\mathcal{N}, \mathcal{P}, \mathcal{Q}\rangle$, where $\mathcal{N}$ is a DNN, $\mathcal{P}$ is a precondition on the inputs, and $\mathcal{Q}$ is a postcondition on the outputs.
    
    {\bf Output}: \sat{} if $\exists\;x\;|\;\mathcal{P}(x) \wedge \mathcal{Q}(\mathcal{N}(x))$, and \unsat{} otherwise.
\end{definition}

In other words, given a safety requirement expressed as a precondition $\mathcal{P}$ on the input space, and a postcondition $\mathcal{Q}$ on the output space, a formal verification tool should return \sat{} if there exists at least one input point 
$x$ which adheres to $\mathcal{P}(x) \wedge \mathcal{Q}(\mathcal{N}(x))$, and \unsat{} otherwise. Usually, the precondition $\mathcal{P}$ encodes all inputs in our domain of interest, and the postcondition $\mathcal{Q}$ usually encodes the \emph{negation}, of the wanted behavior. Hence, an \unsat{} result indicates that the DNN $N$ \emph{always} behaves correctly, while a \sat{} result indicates that at least one input $x$ triggers a bug. 

\medskip
\noindent
\textbf{DNN Verification Example}. Given the toy DNN  illustrated in Fig.~\ref{fig:toy-net}, and given the property \emph{the DNN always outputs a value strictly smaller than $10$}, i.e., we want to verify that for any input $x=\langle v_1^1, v_1^2\rangle$, the condition $N(x)=v_4^1 < 10$ always holds. To formulate this as a verification query, we select a precondition that imposes no restrictions on the inputs, i.e., $P=(\textit{true})$. We then define $Q=(v_4^1\geq 10)$, which represents the \textit{negation} of the desired property. A sound verifier will eventually answer that this query is satisfiable (\sat), along with a feasible counterexample, such as $x=\langle 2, -1\rangle$, leading to $v_4^1=11 \geq 10$. Therefore, the specified property does not hold for this particular DNN.

In the case of adversarial inputs, the verification engine can encode the precondition $\mathcal{P}$ to include a small $\epsilon$-ball  $B_{\epsilon}$ around an input of interest $\mathbf{z}$, while the postcondition $\mathcal{Q}$ can include a misclassification, i.e., an unwanted decision $N(\mathbf{z})\neq N(\mathbf{x})$ --- and hence returning a slightly perturbed input $\mathbf{z} \in B_{\epsilon}(\mathbf{x})$ which serves as an adversarial input to $N$. For more details, we refer to previous work on \texttt{Marabou} \citep{katz2019marabou}. In recent years, the formal methods community has put forth a plethora of tools and algorithms \citep{katz2019marabou, wang2021beta, CountingProVe, wang2018efficient, marzari2023enumerating}. However, the application of these tools in a deep reinforcement learning context is relatively underexplored; although some work has been done in this direction \citep{pore2021safe, marchesini2021benchmarking, corsi2023constrained}, these approaches typically have limited scalability \citep{amir2023verifying, ProVe} and the problem of formalizing the safety requirements in a multi-step context and solving them \emph{efficiently}, still remains an open question \citep{liu2021algorithms}.

\subsection{The \#DNN-Verification Problem}
All sound and complete formal tools eventually answer the aforementioned DNN verification query with a definite \sat{} or \unsat{} answer. Although \unsat{} provides a formal guarantee of safety (i.e., no adversarial input exists), in many cases, a simple \sat{} answer is not informative enough to understand the behavior of the neural network and the level of its susceptibility. For example, it is unclear whether the DRL agents \emph{almost always} violate the required property, or rather, if these violations are extremely rare edge cases. This information is also key in order to improve the trained models against adversarial inputs, as we further elaborate in Sec.~ref{sec:analysis} and Sec.~\ref{sec:retraining}.

In the recent work of \cite{CountingProVe}, the authors formalize, for the first time, an extension of the basic DNN verification problem to the counting version, i.e., the problem of quantifying the number of such violation points in the input space, which cause the DNN to err. Notice that, in the context of DNN, the input space is typically continuous and thus the number of violation points should be viewed as the portion of the input space that includes violating inputs w.r.t. the given specification. The first attempt in this direction has been presented in our previous work, where we presented \prove{}, a tool for over-approximating the unsafe region of a given DNN agent, input domain, and property of interest \citep{ProVe}. \prove{} normalizes the subdomains which include violating inputs, and returns the \texttt{Violation Rate} --- the size of this unsafe portion.
Following Definition~\ref{def:decision_problem}, given the tuple $\mathcal{R}$ we define $\Gamma(\mathcal{R})$ as the set of all the input points that satisfy the conditions specified by the property of interest on the given DNN ($\mathcal{N}$), formally,
\[ \Gamma(\mathcal{R}) = \Bigg\{ x \; \big\vert \; \mathcal{P}(x) \wedge \mathcal{Q}(\mathcal{N}(x)) \Bigg\} \] The solution to the \textit{\#DNN-Verification} problems is then the cardinality of $\Gamma(\mathcal{R})$; formally:

\begin{definition}[\textit{The \#DNN-Verification Problem}]
\label{def:counting_problem}
\phantom{a}

{\bf Input}: A tuple $\mathcal{R}=\langle\mathcal{N}, \mathcal{P}, \mathcal{Q}\rangle$, where $\mathcal{N}$ is a trained DNN, $\mathcal{P}$ is a precondition on the input, and $\mathcal{Q}$ a postcondition on the output.

{\bf Output}: 
$\vert \Gamma(\mathcal{R})\vert$
\end{definition}

\noindent
\textbf{Complexity.}
A well-known limitation of formal verification is its scalability with respect to the size of the DNN in question. It has already been shown in previous work that the DNN verification problem is NP-complete \citep{KaBaDiJuKo17}, and thus, even verifying small DNNs can become challenging. This problem is aggravated in the context of counting all \sat{} assignments, as required in the case of the \#DNN verification problem --- which has been proven to be \#P-complete \citep{CountingProVe}. Therefore, in recent years, some attempts have been made to provide an approximate solution. In particular, in the same work, the authors proposed an approximation algorithm that provides a probabilistic bounded solution that computes the results in a fraction of the time.

\section{Analytical Tools and Techniques} 
\label{sec:tools}

In this section, we introduce the tools and algorithms that were employed for our analysis, followed by a detailed description of the two benchmarking environments and details regarding the training procedure we adopted. 

\medskip
\noindent
\textbf{Note.} It is crucial to highlight that our approach remains \emph{agnostic} to the specific training procedure of the model. Consequently, the analysis (detailed in Sec.~\ref{sec:analysis}) is conducted in an \emph{offline} setting, post-training. Nonetheless, in Sec. \ref{sec:retraining}, we discuss some training-driven methods that can effectively mitigate the vulnerabilities discussed in our analysis.

\subsection*{ProVe: A Formal Tool for Verification, Counting, and Enumeration of Adversarial Inputs}

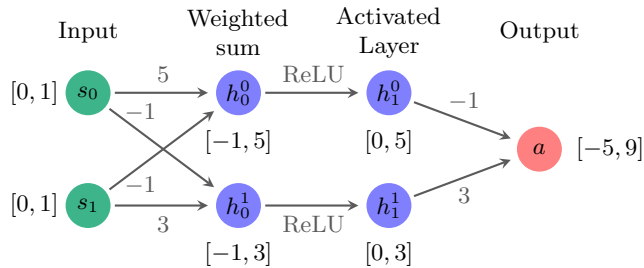
\begin{figure}[t]
    \begin{center}
    \def\layersep{2.0cm}
    \begin{tikzpicture}[shorten >=1pt,->,draw=black!50, node
    distance=\layersep,font=\footnotesize]
    
    \node[input neuron] (I-1) at (0,-1) {$s_0$};
    \node[input neuron] (I-2) at (0,-2.5) {$s_1$};
    
    \node[left=-0.05cm of I-1] (b1) {$[0, 1]$};
    \node[left=-0.05cm of I-2] (b2) {$[0, 1]$};
    \node[hidden neuron] (H-1) at (\layersep,-1) {$h^0_0$};
    \node[hidden neuron] (H-2) at (\layersep,-2.5) {$h^1_0$};	
    \node[hidden neuron] (H-3) at (2*\layersep,-1) {$h^0_1$};
    \node[hidden neuron] (H-4) at (2*\layersep,-2.5) {$h^1_1$};
    \node[output neuron] at (3*\layersep, -1.75) (O-1) {$a$};
    
    \draw[nnedge] (I-1) --node[above] {$5$} (H-1);
    \draw[nnedge] (I-1) --node[above, pos=0.3] {$-1$} (H-2);
    \draw[nnedge] (I-2) --node[below, pos=0.3] {$-1$} (H-1);
    \draw[nnedge] (I-2) --node[below] {$3$} (H-2);
    \draw[nnedge] (H-1) --node[above] {ReLU} (H-3);
    \draw[nnedge] (H-2) --node[below] {ReLU} (H-4);
    \draw[nnedge] (H-3) --node[above] {$-1$} (O-1);
    \draw[nnedge] (H-4) --node[below] {$3$} (O-1);
    
    \node[below=0.05cm of H-1] (b1) {$[-1, 5]$};
    \node[below=0.05cm of H-2] (b2) {$[-1, 3]$};
    
    \node[below=0.05cm of H-3] (b1) {$[0, 5]$};
    \node[below=0.05cm of H-4] (b2) {$[0, 3]$};
    \node[right=0.05cm of O-1] (b1) {$[-5, 9]$};
    
    \node[annot,above of=H-1, node distance=0.8cm] (hl1) {Weighted sum};
    \node[annot,above of=H-3, node distance=0.8cm] (hl2) {Activated Layer };
    \node[annot,left of=hl1] {Input };
    \node[annot,right of=hl2] {Output };
    \end{tikzpicture}
    \caption{An example of interval propagation for a reachability approach to verification.}
    \label{fig:tools:propagation-example}
    \end{center}
\end{figure}

In our analysis, the primary and foundational tool we employ is \prove{}. This tool, when provided with a set of safety requirements, can effectively partition the input domain into two types of regions. The first is a safe portion where the specified safety properties are upheld, while the second is an unsafe portion where at least one of the given requirements is violated. We originally presented this approach in \cite{ProVe}, where we also introduced a novel metric known as the \adversarialrate{} (or \texttt{Violation Rate}). The \adversarialrate{} is computed \textcolor{black}{as the percentage of the input domain that is covered by unsafe inputs, i.e., the size of the unsafe region expressed as a percentage of the total input domain size. The metric therefore ranges from $0\%$ (no adversarial inputs exist) to $100\%$ (the entire input domain is adversarial).} \prove{} is built upon the fundamental idea of \textit{interval propagation}. Illustrated in Fig.~\ref{fig:tools:propagation-example} is an instance of this concept, where the objective is to formally verify that if the input vector ${\bf v} = (s_0, s_1)$ belongs to the interval $[[0, 1], [0, 1]]$, the network should consistently produce an output value that is lower or equal to $10$. When successively propagating the input interval layer-by-layer, we derive an abstraction (or overestimation) of the reachable output set. In this specific example, we determine that $a \in [-5, 9]$, with the upper bound of $9$ satisfies the stipulated requirement. Consequently, we formally establish that the DNN adheres to the specified criteria.

While the presented example may seem straightforward, it is crucial to note two significant challenges: (i) the process yields an \textit{overestimation}, with the precision deteriorating with the size of the propagated input interval, and (ii) the approach is incomplete, i.e., it may not find a concrete counterexample, even when it exists, and hence, it may fail to provide an answer. For instance, consider the scenario depicted in Fig.~\ref{fig:tools:propagation-example} where the aim is to prove that the output is always less than $5$. In such cases, this approach falls short as (based on simple interval arithmetic) the requested value belongs to the interval $[-5, 9]$. To overcome this limitation, \prove{} exploits an additional technique: \textit{iterative splitting} (a.k.a. \textit{iterative refinement}). The intuition is to iteratively split the input domain into a set of subdomains and analyze them independently. This approach poses two main advantages. Firstly, it reduces the overestimation --- this is because feed-forward DNNs are Lipschitz-continuous \citep{wang2018efficient}, and hence their error is inversely proportional to the interval width. Secondly, it allows us to estimate the shape of the output functions, and thus to directly compare the different output nodes (or an output interval and a reference value, as is the case in the previous example). 

\begin{figure}[t]
\centering
\minipage{0.3\linewidth}
    \includegraphics[width=\linewidth]{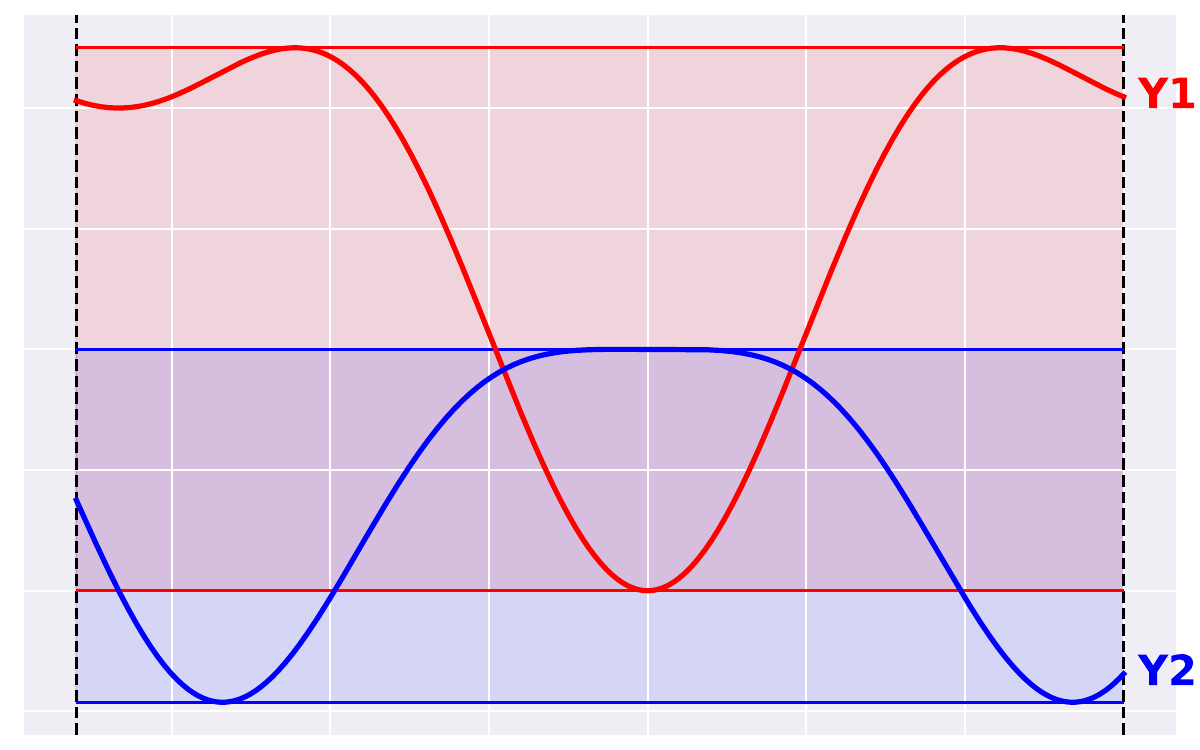}
\endminipage \hfill
\minipage{0.3\linewidth}
    \includegraphics[width=\linewidth]{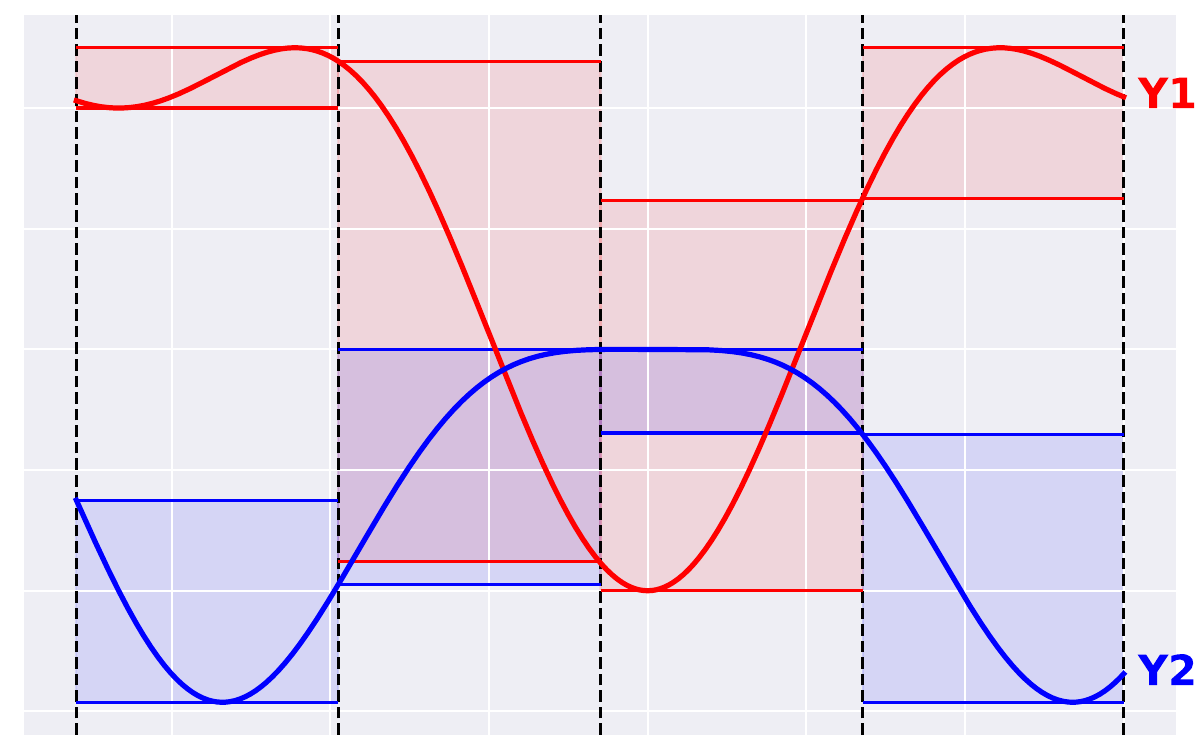}
\endminipage \hfill
\minipage{0.3\linewidth}
    \includegraphics[width=\linewidth]{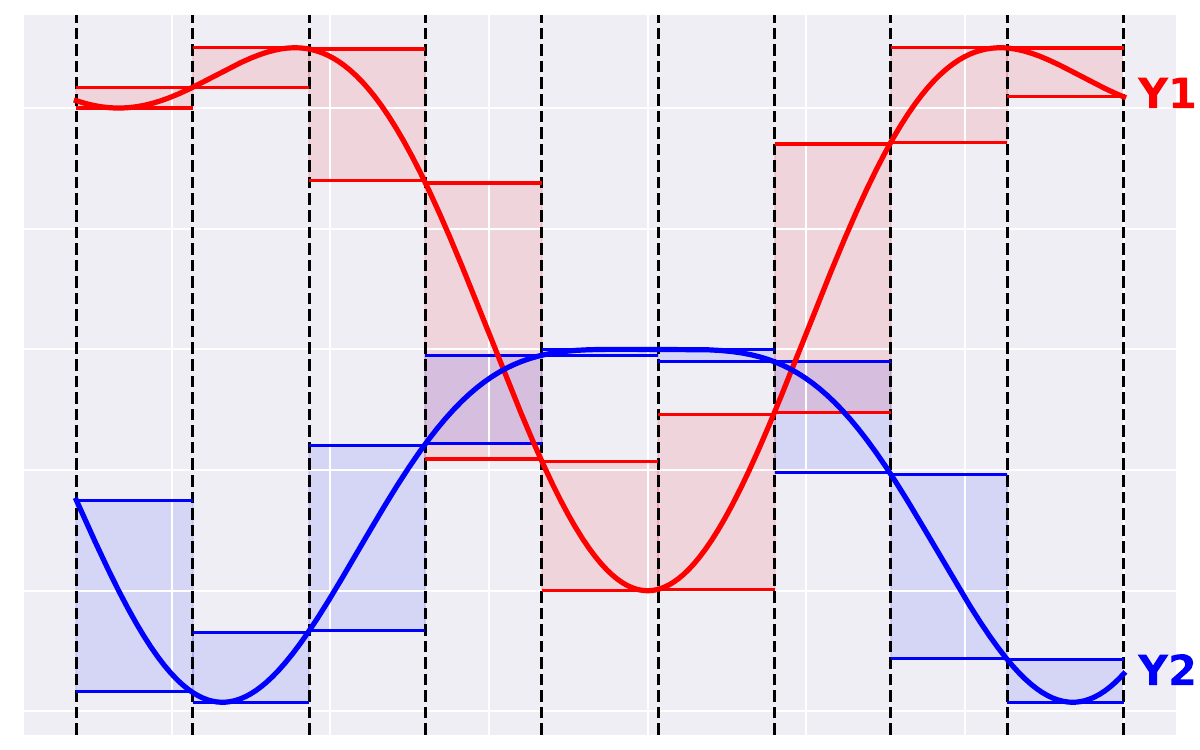}
\endminipage
\caption{A toy example for the iterative splitting procedure of \prove{}. In the case depicted in the first figure, it is not possible to formally prove where $Y1$ is greater than $Y2$ given that the upper and lower bounds overlap. In the second and third figures, the iterative splitting procedure allows the division of the input domain into safe and unsafe regions.}
\label{fig:tools:iterative-splitting}
\end{figure}

Fig.~\ref{fig:tools:iterative-splitting} illustrate an example of this approach. Suppose we wish to prove that one output ($y_1$) is always greater than the other ($y_0$); the leftmost figure depicts a scenario in which it is impossible to provide a formal answer as the propagated output reachable sets overlap, while in the rightmost figure, it is possible (by exploiting the input splitting) to formally show where the output $y_1$ is greater than $y_0$. Finally, to compute the \adversarialrate{}, \textcolor{black}{\prove{} sums the sizes of the sub-intervals that violate the requirement and expresses the result as a percentage of the size of the initial domain.} Notice that, it is possible that for some sub-intervals the reachable sets still overlap. In such cases, \prove{} iteratively performs additional domain splits until it can return an answer or reach an $\epsilon$ precision factor. \prove{} is sound and $\epsilon$-complete; we refer the reader to the original paper for the completeness proof and the complexity analysis \citep{ProVe}. Crucially, for the single interval propagation, \prove{} can rely on different verification tools as a backend, such as \texttt{Neurify} \citep{wang2018efficient}. Notice that, a complete version of \prove{} has been presented in \cite{CountingProVe} based on \texttt{Marabou}  \citep{katz2019marabou} --- a sound and complete verification engine used for various verification use cases \citep{BaKa23, AmScKa21, AmWuBaKa21, AmZeKaSc22, AmFrKaMaRe23, AmMaZeKaSc23, CaKoDaKoKaAmRe22}.

\medskip
\noindent
\textbf{Code.}
The code for \prove{} is publicly available at: \url{https://github.com/d-corsi/NetworkVerifier}.

\subsection*{Counting ProVe: Finding an Approximated Solution}

Although \prove{} stands out as a potent tool for \#DNN-Verification, its soundness, and $\epsilon$-completeness introduce significant scalability challenges \citep{ProVe}. The drawback of formal methods-based tools has prompted numerous efforts to enhance efficiency through the utilization of diverse algorithmic optimizations \citep{bunel2020branch}, meticulously crafted heuristics \citep{liu2021algorithms} and GPU acceleration \citep{wang2021beta}. However, in some cases, the problem can remain intractable, hence requiring the employment of approximation techniques. In this context, \cite{CountingProVe} propose \countingprove{}, a solver that allows efficiently approximating a solution, crucially delivering a set of probabilistic guarantees regarding the quality of the obtained estimation.

The key intuition behind this approach is as follows. 
By iteratively partitioning the input domain, \countingprove{} generates an assignment tree in which each node corresponds to a subspace of the initial input domain. If we assume that we can split the number of adversarial inputs evenly into the two sub-spaces each time we split the input domain, then for the computation of the number of violation points in the entire input space, it would be sufficient to count the total violation points in the subspace of a single leaf (which is easy to compute) and then multiply it by $2^s$ (where $s$ is the number of splits). However, since it is not possible to guarantee a perfectly balanced distribution of the violation points at each domain split, we propose to use a heuristic-based strategy. Interestingly, however, the authors formally show that independent of the chosen heuristic, they can provide strong probabilistic guarantees regarding the quality of the result. We refer to \cite{CountingProVe} for additional details and the full proofs. 

Empirically, \countingprove{} has shown a good tradeoff between the accuracy of the results and the computational time required to compute the \adversarialrate{}, especially when the size of the unsafe region is large. However, it struggles to provide a good estimation when this value is particularly low (i.e., approximately $\leq 1\%$); in such cases, the best solution is to exploit a formal verifier if we are interested in providing a formal guarantee of a zero-violation condition. For our analysis, we exploit \countingprove{} for a preliminary evaluation, however, in the cases where it returns a \adversarialrate{} of less than $1\%$, we perform an additional formal and complete analysis with \prove{} (or \texttt{Marabou}) to confirm the quality of the obtained results.

\subsection*{Proximal Policy Optimization and Twin Delayed DDPG}

In the upcoming sections, we outline two cutting-edge approaches for training DRL policies: PPO and TD3. It is important to highlight that our discussion of these training algorithms remains at a high level, as the subject matter of this work is on harnessing formal methods, where the primary emphasis is on verifying DRL policies \emph{post-training}. However, we also touch upon several training techniques aimed at addressing adversarial inputs in Section~\ref{sec:retraining}.

In general, DRL is a popular paradigm in Machine Learning, in which an RL-trained agent is realized with a DNN. Unlike the case in supervised learning, DRL does not rely on a dataset, but rather the agent learns through a trial-and-error process, in which it iteratively attempts to maximize its overall \emph{reward} function.
More formally, the objective of a DRL agent is to learn a \emph{policy} $\pi$, which establishes a mapping from a \emph{state} $s$ of the environment to an appropriate \emph{action} $a$. The nature and interpretation of $\pi$ can vary among different learning algorithms. In some contexts, typically \textit{policy gradient} approaches \citep{sutton2018reinforcement, schulman2015trust}, $\pi$ represents a probability distribution over the action space;  while in \emph{value-based} approaches, we try to estimate a desirability score (a.k.a. Q-function \cite{mnih2013playing}) over the possible future actions, and the policy $\pi$ becomes to follow the highest value. Other approaches, such as DDPG \citep{lillicrap2015continuous} and its improved variations (e.g., TD3 \citep{fujimoto2018addressing} and SAC \citep{haarnoja2018soft}) attempt to combine both of the aforementioned methods.

In our evaluation (see Sec.~\ref{sec:analysis}), we primarily utilize the widely recognized Proximal Policy Optimization (PPO) algorithm \citep{ShWoDh17}. This choice is driven by PPO's reputation as the most robust end-to-end DRL algorithm, leading to its extensive adoption in cutting-edge DRL platforms applied in real-world products such as ChatGPT4 \citep{openai2023gpt4} and Sony GT5 \citep{wurman2022outracing}. Given its outstanding performance, PPO is also the predominant algorithm for training DRL agents in the field of robotics and control \citep{akkaya2019solving, ray2019benchmarking}, which is the focal point of our study. In our experimental assessment, we incorporate also the TD3 algorithm as it is also a widespread approach for DRL training, and is a competing value-based alternative to PPO\footnote{We note that in the case of actor-critic algorithms, the taxonomy does not have a unanimous consensus; in this paper, we follow the idea that DDPG mainly relies a Q-function estimation and is thus considered an off-policy and value-based approach.}. 
The decision to perform the analysis on two algorithms with a completely different approach stems from our objective to illustrate that the occurrence of adversarial examples does not directly depend on the algorithm of choice but is more likely to be an inherent property of the neural networks in question. Nevertheless, we recognize that recent work proposes algorithmic improvements that increase the performance of the trained agents \citep{van2016deep, marchesini2023improving, schulman2015high}. 


\section{Benchmarking Environments}
\label{sec:benchmarks}
In this section,  we present the two benchmarking environments used for our analysis. For each of them, we will briefly introduce the task, highlight various details regarding the training phase (e.g., reward function and network topology), and discuss the safety requirements we aim to guarantee. Notice that, in some of our analyses, we alter the structure of the neural network to evaluate whether this affects the presence and impact of the adversarial inputs, however, in this section, we report the \textit{basic setup} for the training that is shared among all the experiments.

The first environment is a customized version of standard GridWorld that we call \gridenv{}, which is our running example. This choice is motivated mainly because the environment is fully observable and low-dimensional --- allowing us to encode a set of safety requirements that covers \texttt{all} possible unsafe actions; implying that if an agent has a $0\%$ \adversarialrate{} it can not collide. Moreover, the input features of the network hold a spatial interpretation, which simplifies the graphical representation and helps in understanding the results. Our second benchmark is the \turtleenv{}, which is a real-world robotic benchmark and a well-known task in the DRL literature \cite{marchesini2021benchmarking, tai2017virtual}. \turtleenvshort{} will serve as a proof of concept, to show that our findings are replicated in more complex environments.

\subsection{Jumping World}
\begin{figure}[t]
\centering
\minipage{0.18\linewidth}
    \includegraphics[width=\linewidth]{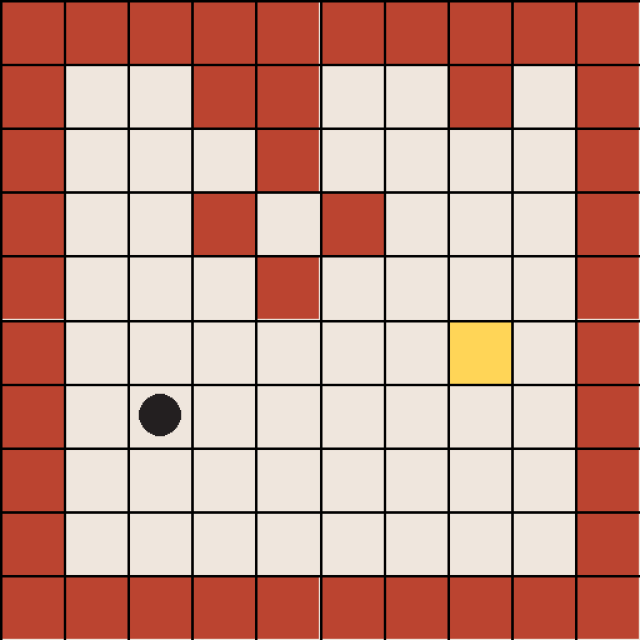}
\endminipage \hfill
\minipage{0.18\linewidth}
    \includegraphics[width=\linewidth]{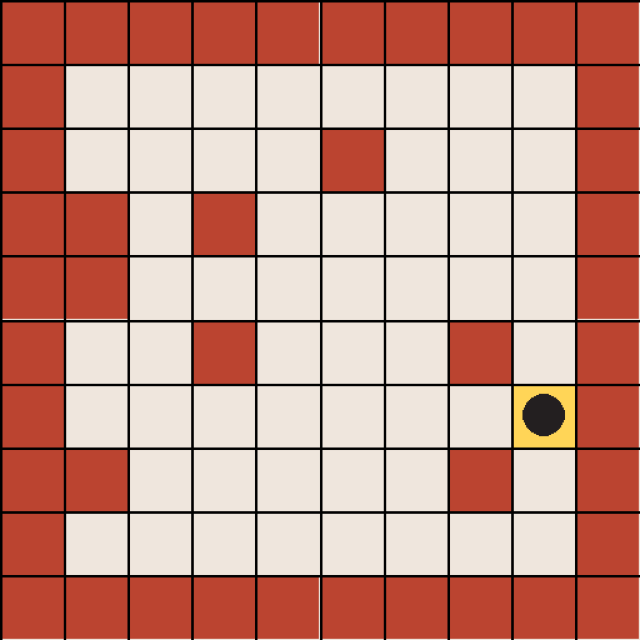}
\endminipage \hfill
\minipage{0.3\linewidth}
    \includegraphics[width=\linewidth]{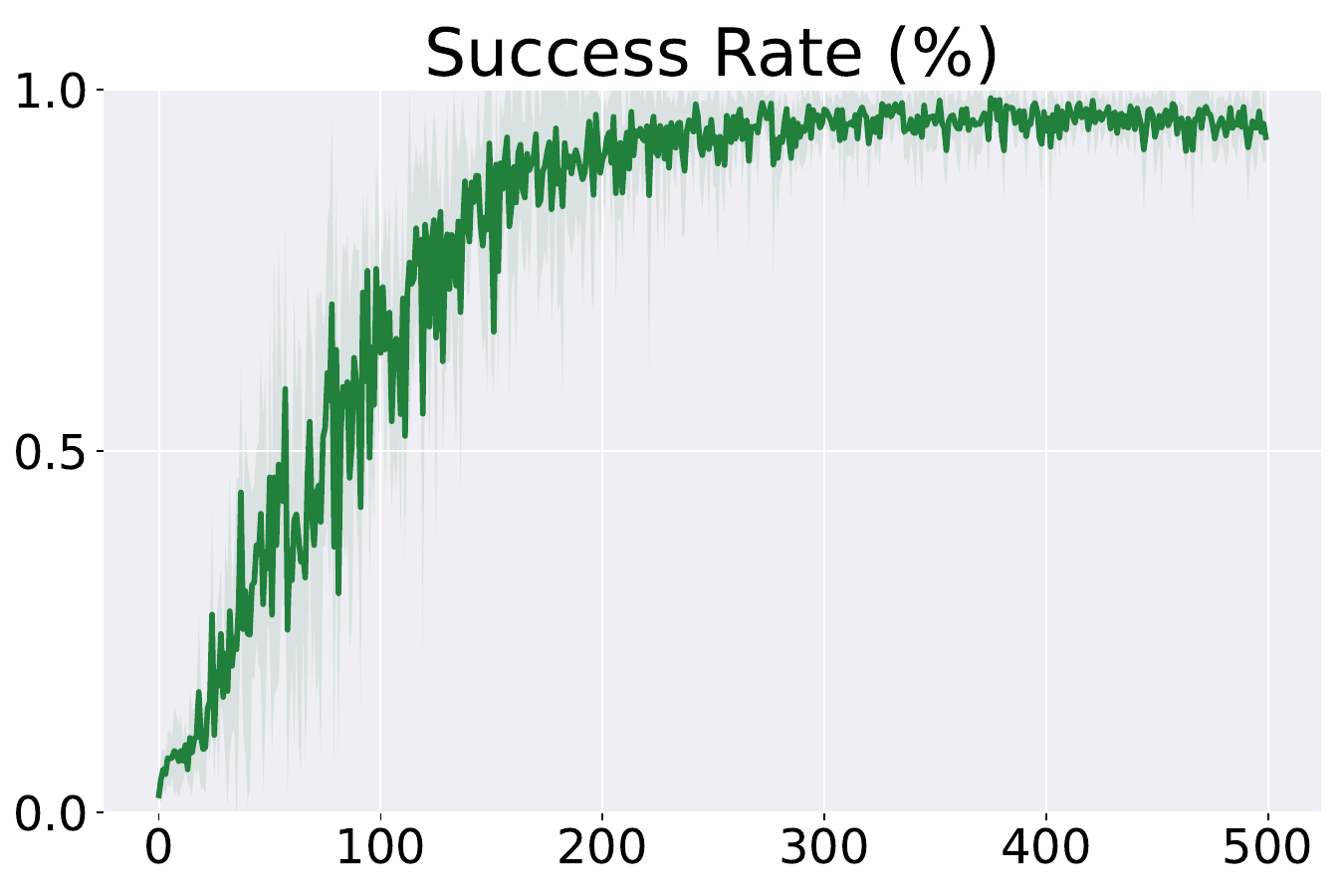}
\endminipage \hfill
\minipage{0.3\linewidth}
    \includegraphics[width=\linewidth]{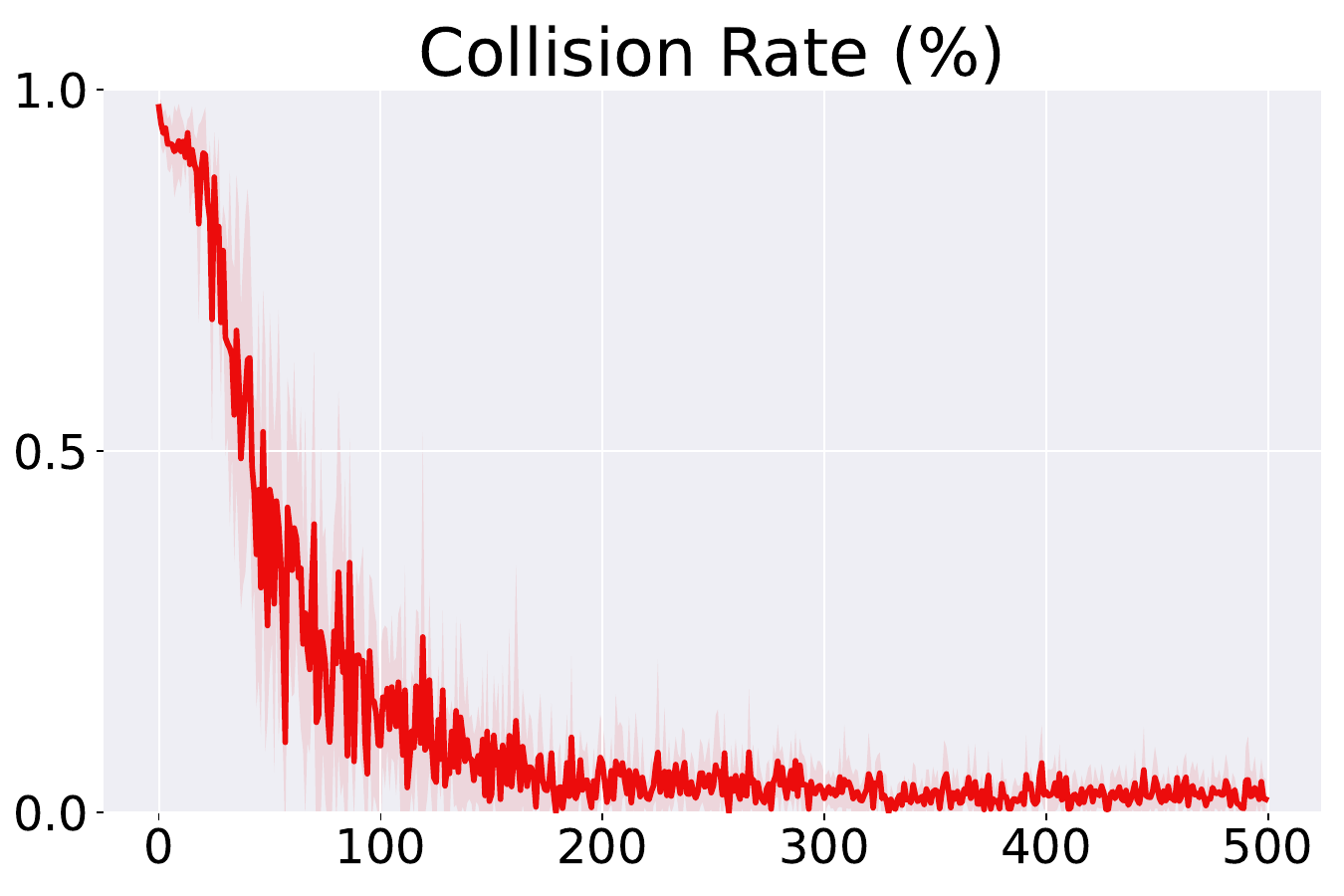}
\endminipage
\caption{The \gridenv{} environment analyzed for our experimental evaluation. On the left is a screenshot from the simulation and on the right are the empirical results of our training phase.}
\label{fig:environments:gridworld}
\end{figure}

Our first environment is depicted in Fig.~\ref{fig:environments:gridworld}. The goal of the agent (black dot) is to reach the target position (yellow square) while avoiding collisions with the obstacles (red squares) that are randomly generated at each new episode. This environment is an abstraction of a real-world navigation problem. The agent is equipped with $4$ proximity sensors that detect whether there is an obstacle in the corresponding direction. For explanatory purposes, we designed the agent to respect two strict requirements: (i) the state space should be continuous, to allow infinite-many inputs, and hence, the potential presence of adversarial inputs; 
and (ii) the consequence of an action should be deterministic, to allow the encoding of a potentially large (but limited) set of safety requirements that, if respected, guarantee that the agent never collides. To respect these two requirements, we designed a specific model of the environment. In particular, the agent moves towards a set of discrete cells, with the addition of Gaussian noise per each state observation. Broadly speaking, this setting allows the agent to ``know'' the current cell, without having precise information about its current position within the continuous space of the cell. Finally, the target position changes over the episodes, and hence we designed the coordinates of the target to also be part of the observation space, and hence an input to the agent.

\textcolor{black}{\noindent\textbf{Remark on continuous state space.} We acknowledge that the addition of Gaussian noise to an otherwise discrete grid is somewhat artificial, and serves primarily as a proof of concept to enable formal analysis in a continuous input domain. In principle, our approach is also applicable to fully discrete environments; however, adversarial inputs in a purely discrete setting are significantly less likely to arise in practice, since any perturbation must align exactly with the finite discretization of the state space to induce a different action.}

\paragraph{State/Action Spaces and DNN Topology.}
The state space of the agent consists of $8$ features. The first two observations represent the position of the agent in the environment, in cartesian coordinates. As previously described, these two readings are noisy and hence match our continuous scenario. More precisely, there is a random noise that ranges from $-0.5$ to $0.5$ on both axes. \textcolor{black}{Crucially, the perturbation radius used in the formal verification queries is set to match exactly the size of one grid cell (i.e., the full noise range of $[-0.5, 0.5]$ per axis). This design choice ensures that the distribution of adversarial inputs can be directly represented and visualized in correspondence with the grid layout of the arena, as detailed in Sec.~\ref{sec:analysis}.} The next four observations represent the reading of the obstacle sensors, and thus each reading includes a binary value indicating if there is an obstacle in the corresponding direction. More formally, $S_{0,1} \in \mathcal{R}^2$ and $S_{2,3,4,5} \in \langle 0, 1\rangle$. Finally, the last two features represent the coordinates of the target. The action space is discrete with $4$ possible actions \textit{left}, \textit{right}, \textit{up}, and \textit{down}. As previously mentioned, the consequence of an action is deterministic, i.e., the agent always moves to the desired cell but does not know the exact position inside it. The agents trained on this task was realized by a neural network with a feed-forward architecture, with $8$ inputs and $4$ outputs, respectively corresponding to the state and action spaces. Finally, the DNN consists of two additional hidden layers of $32$ neurons with ReLU activations.

\paragraph{Training Details.} We trained our agent with the PPO algorithm. Solving this task is not particularly challenging for a state-of-the-art DRL algorithm and thus we relied on the standard settings. During training, we employed the following (discrete) reward function:
\begin{equation}
    r_t =
    \begin{cases}
    1 & \text{if the goal is reached} \\
    -1 & \text{if the agent collides} \\
    -0.01 & \text{otherwise}
    \end{cases}
\end{equation}
\noindent notice that the two main components of the reward function (i.e., $\pm1$ are for terminal states) and a small penalty of $-0.01$ for each time-step. The penalty encourages the agent to find the shortest path to the target. \textcolor{black}{As is standard in DRL, the agent is trained to maximize the expected discounted cumulative reward $R_t = \mathbb{E}\big[\sum_{t} \gamma^{t} \cdot r_t\big]$, where $\gamma \in [0,1]$ is the discount factor (see Sec.~\ref{sec:background}).}

\paragraph{Safety Requirements.} The \gridenv{} environment is an abstraction of a real-world navigation problem. Hence, the natural safety property to verify (or find violations thereof) is collision avoidance. To better define the safety requirements, we recall the definition of Sec. \ref{sec:background}, in this environment the first two features of the observation space represent the current position of the agent (i.e., $S_0$ and $S_1$) are not relevant for the safety, for this reason, the safety properties can be formulated as follows: \textit{``if the agent identifies an obstacle in its proximity, then the agent must not move towards that direction in the next time-step"}.
\textcolor{black}{Although this property may appear trivial, it is specifically designed to demonstrate that even such elementary safety constraints can be violated by a well-trained DRL agent when subjected to small input perturbations. The key insight is that formal verification reveals susceptibilities that standard empirical evaluation routinely misses, even for requirements that a rational agent should trivially respect.}

\subsection{Robotic Mapless Navigation}
\begin{figure}[t]
\centering
\minipage{0.28\linewidth}
    \includegraphics[width=\linewidth]{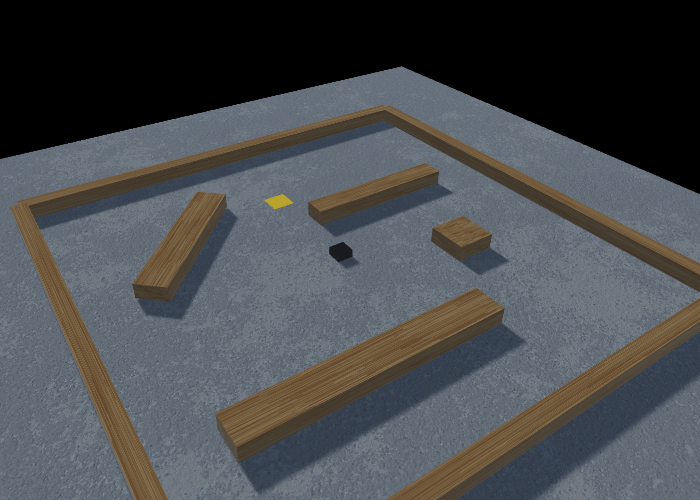}
\endminipage \hfill
\minipage{0.34\linewidth}
    \includegraphics[width=\linewidth]{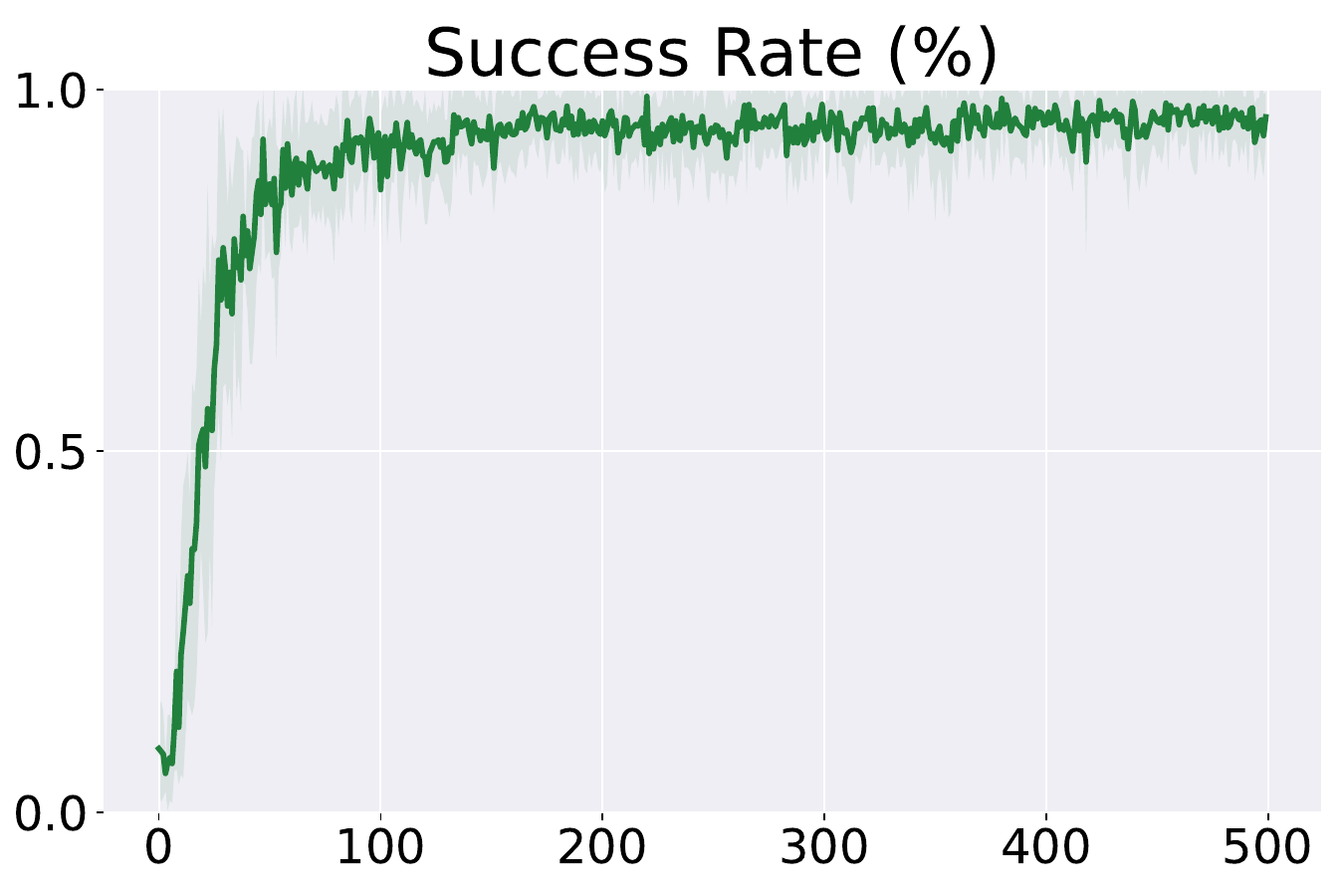}
\endminipage \hfill
\minipage{0.34\linewidth}
    \includegraphics[width=\linewidth]{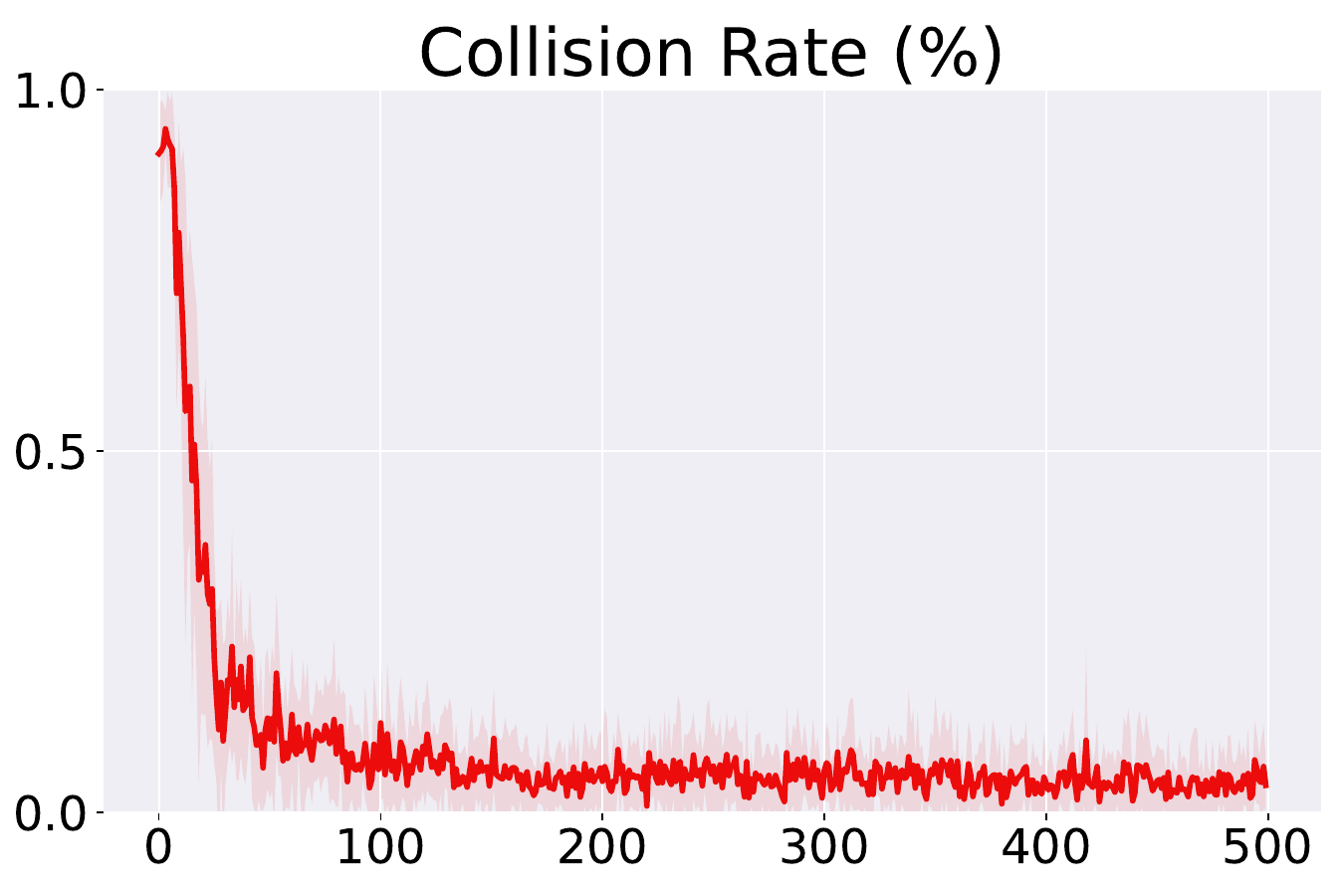}
\endminipage
\caption{The \turtleenv{} environments analyzed for our experimental evaluation. On the left is a screenshot from the simulation and on the right are the empirical results of our training phase.}
\label{fig:environments:turtlebot}
\end{figure}

The second environment is depicted in Fig. \ref{fig:environments:turtlebot}. \turtleenv{} is a classic benchmark in the DRL community \citep{tai2017virtual, marchesini2020discrete} and is considered challenging even for state-of-the-art algorithms. Robotic navigation is the task of navigating a robot (in our case, \emph{TurtleBot4} \citep{thale2020ros}) in an arena, while avoiding possible obstacles and, in some cases, adhering to additional requirements. In this work, we focus on a more challenging variant of the problem, called \textit{mapless navigation}, where the agent does not have access to a global map of the environment and can rely only on its local sensor readings. 
In our setup, we rely on the recent work from \cite{marchesini2020discrete} and \cite{amir2023verifying}, where the observation space of the robot includes the readings from a lidar sensor (i.e., the distance from the closest obstacle in the given direction), and the relative polar coordinates of the target (i.e., angular distance). However, our setup is slightly different with respect to the previously cited work as we rely on a continuous action space (i.e., linear and angular velocity) following the control strategy suggested in the work of \cite{tai2017virtual}. This design choice is motivated by three main reasons: (i) it allows us to showcase that our tools can perform the analysis on the adversarial input also on a continuous action space; (ii) it represents a more realistic scenario, where a robot should typically be controlled with continuous actions; and (iii) it allows to directly execute the neural controller on the real-world robot to validate our findings.

\paragraph{State/Action Spaces and DNN Topology.} The state space of the environment consists of $17$ features. The first $15$ observations represent the lidar readings, from left to right, with a step of $\approx 13^\circ$, normalized between $0$ and $1$, where $1$ means that no obstacle is detected in the corresponding direction and a $0$ value indicates that the robot collided into an obstacle at the current step. The other two observations represent the polar coordinates of the target cell (also, normalized in the range $[0, 1]$). The action space consists of two actions, both having a continuous value in the range $[-1, 1]$. The two actions encode the adjusted linear velocity and angular velocity of the robot. We refer to our previous work on this topic \citep{amir2023verifying} for additional details on the network setup and topology. For this problem, we employ a feed-forward neural network with $17$ inputs and $2$ outputs. The network includes two hidden layers of $64$ neurons and ReLU activations.

\paragraph{Training Details.}
Fig. \ref{fig:environments:turtlebot} presents the obtained results from our training phase. Unlike the previous \gridenv{} environment, and in accordance with the literature \citep{tai2017virtual, marchesini2022enhancing}, it is quite non-trivial to obtain a model with more than $90\%$ success rate on this complex task. We thus impose a $95\%$ cutoff to consider a model \textit{successful}. For this task, we employ a continuous reward function:
\begin{equation}
    R_t = \begin{cases}
    1 & \text{if the goal is reached} \\
    -1 & \text{if the agent collides} \\
    (dist_{t-1} - dist_{t}) \cdot \eta - \beta & \text{otherwise}\\
    \end{cases}
\label{eq:methods:reward_function}
\end{equation}

\noindent where $dist_k$ is the distance from the target at time-step $k$; $\eta$ is a normalization factor; and $\beta$ is a penalty, intended to encourage the robot to reach the target quickly (in our experiments, we empirically set $\eta=3$ and $\beta=0.001$). \textcolor{black}{As in the \gridenv{} task, the agent is trained to maximize the expected discounted cumulative reward with a discount factor $\gamma \in [0,1]$, following the standard DRL setting.} We trained our basic models with the PPO algorithm and the average learning curve we obtained is presented in Fig. \ref{fig:environments:turtlebot}.

\paragraph{Safety Requirements.}
This benchmark is a classical robotic navigation problem and thus the crucial requirement is trivially collision avoidance. However, our approach can also include additional safety properties, e.g., \textit{do not move in a specific direction if the lidar sensor detects an obstacle in the corresponding direction}, etc. Unlike in the \gridenv{} environment, there is not a clear correspondence between an action and its consequences, and this is due to the partial observability of the \turtleenv{} environment and the continuous action spaces. For this reason, we rely on a set of finite properties that were covered in previous studies \citep{ProVe, amir2023verifying}. In such a context, our goal is not to cover all the possible collisions, but to prove that a given model is completely safe with regard to the given requirements.
\section{Analysis}
\label{sec:analysis}

In this section, we present a comprehensive analysis of several characteristics related to adversarial inputs among DRL agents. Specifically, we show that this problem can generate unsafe behaviors even among models that appear completely safe after an exhaustive empirical evaluation. In addition, we perform a series of studies to better understand the behavior of adversarial configurations from both a spatial and temporal perspective. Finally, we try to understand if specific neural network configurations can mitigate the vulnerability to adversarial perturbations, in various settings.

We analyzed \emph{thousands} of agents from two popular DRL benchmarks: (i) \gridenv{}, and (ii) \turtleenv{}. In particular, we perform the initial qualitative analysis on the \gridenv{} benchmark. The observability and limited state space of this environment allow us, under some circumstances that we later elaborate on, to formally prove whether an agent trained on this task is completely safe. Hence, by designing finite (but, possibly many) properties, we can cover all possible cases in which an agent may potentially exhibit unwanted behavior, and verify its safety, or lack thereof. Crucially, due to the geographical interpretation of the input features (as discussed in the previous section), it is easier to visualize all the states in which the agent may potentially act in an unsafe manner, providing a spatial intuition regarding the frequency of the adversarial inputs in the state space (e.g., Fig. \ref{fig:analysis:spatial}). We also apply our analysis on the \turtleenv{} benchmark, in which an agent learns to control and navigate \emph{TurtleBot 4} --- a real-world robotic navigation platform, that operates in an \emph{unknown} arena.

In each of the following subsections, we first present an open question related to DRL safety, describing the experimental setup for the analysis and the tools involved in tackling the question. Then, we present our results per each question of interest. This is done by quantitative and qualitative analysis of formal verification queries, on both of the aforementioned benchmarks. When possible, we also relate the observed behaviors to previous research and discuss the possible implications of our results in a general, and more wholesome, manner.

\subsection{Identifying Adversarial Inputs: Random Search vs. Verification}

In our first evaluation, we show that it can be surprisingly challenging to: (i) identify the existence of adversarial inputs, and (ii) quantify the rate at which such adversarial inputs exist. 

\medskip
\noindent
\textbf{Training.}
To demonstrate this, we first trained $n=25$ models w.r.t. the \gridenv{} benchmark. These were generated by choosing $k=5$ seeds, each giving rise to $5$ different agents --- with each agent being a single instance of a policy at a given moment throughout the training process. We tested each of the models and empirically validated them for $500$ episodes, each episode including between $200$ and $300$ time-steps, resulting in a total of approximately $120,000$ (potentially dangerous) actions per model. We selected only the models that achieved a \successrate{} of, at least, $95\%$; this is important to show that any adversarial inputs are not simply due to insufficient training bringing to poor generalization.

\medskip
\noindent
\textbf{Evaluation.}
After training, we set out to identify the adversarial inputs, i.e., inputs that cause collisions in our case (as depicted in Fig.~\ref{fig:environments:gridworld}, we have a collision if the agent moves in a cell occupied by an obstacle). As a first step, we calculated the number of collisions identified in the long training phase, per each agent. We refer to this as the ``standard" \collisionrate{} because it is computed by randomly sampling from the entire state space, without any prior assumptions. All $n=25$ selected agents did not collide in any of the episodes, effectively bringing their \collisionrate{} to $0$. 

Next, we searched again for adversarial inputs. However, unlike the case for the \collisionrate, the second analysis was not conducted by the random state in which the agent is, but rather --- strategically searching the input space with formal methods. 
This was done in the following manner: 
(i) we executed \prove{} on the safety properties defined in Sec. \ref{sec:tools} to formally verify the presence (or lack thereof) of adversarial inputs; 
(ii) if \prove{} returned \unsat, we updated the \adversarialrate{} to be $0$ as we are formally guaranteed that no such adversarial input exists; otherwise 
(iii) \adversarialrate{} will be strictly larger than $0$, in which case we will sample points in the proximity of the counterexamples that \prove{} returns. \textcolor{black}{This motivates the introduction of the \textbf{Adversarial Collision Rate} (\fvcollisionrate{}): an \emph{empirical} metric defined as the collision rate observed when the agent is initialized and executed from states sampled in the proximity of the adversarial inputs identified by the verification tool. Unlike the \adversarialrate{}, which is computed formally by \prove{} over the input domain, the \fvcollisionrate{} is obtained by running the agent in simulation and measuring how frequently it collides when starting near the formally identified violating inputs.}

Our results (summarized in Table~\ref{tab:analysis:adv-testing}) showcase the merits of employing verification-driven techniques to identify adversarial inputs. First, it can be observed that even after extensive testing, all models are susceptible, in varying levels, to adversarial inputs, as indicated by the strictly positive values in the \adversarialrate{} column, which was calculated using \prove. Furthermore, we performed an additional empirical evaluation in the proximity of the detected violating points, showing that the \adversarialrate{} drastically increases. This finding confirms that the considered agents were not safe to deploy, although the empirical evaluation did not identify any cases in which these collisions occurred.

\begin{table}[t]
\centering
\begin{tabular}{|c|c|c|c||c|}
\hline
Seed & \successrate & \collisionrate & \adversarialrate & \fvcollisionrate{} \\
\hline
0 & $92\pm0.3$ & $0\pm0.0$ & 0.09 & $2.1\pm0.1$ \\
1 & $95\pm0.4$ & $0\pm0.0$ & 0.05 & $1.8\pm0.3$ \\
2 & $91\pm0.8$ & $0\pm0.0$ & 0.09 & $2.1\pm0.1$ \\
3 & $93\pm0.2$ & $0\pm0.0$ & 0.07 & $2.0\pm0.1$ \\
4 & $94\pm0.4$ & $0\pm0.0$ & 0.06 & $1.1\pm0.7$ \\
\hline
\end{tabular}
\caption{\gridenv{}: A comparison between the different seeds. Note that the remaining part that is neither a collision nor a successful run is due to the presence of infinite loops, which are not considered safety violations.}
\label{tab:analysis:adv-testing}
\end{table}

\medskip
\noindent
\textbf{Implications.} 
The first implication of this evaluation, as realized by comparing the \collisionrate{} and the \adversarialrate{}, is that formal verification can efficiently identify adversarial inputs that random sampling cannot, even when done extensively. 
Another implication is that by sampling the proximity of points with violations (identified by the \adversarialrate{}) we are able to find additional safety violations (as expressed by the \fvcollisionrate{}). 

\subsection{Spatial Analysis of Adversarial Inputs}
The previous results motivate the need for a deeper understanding of the characteristics of adversarial inputs in the context of DRL. Specifically, the high concentration of adversarial inputs in certain parts of the input domain (as observed in the correlation of \adversarialrate{} and \fvcollisionrate{} in Table~\ref{tab:analysis:adv-testing}), along with the fact that none of these adversarial inputs were found in extensive sampling (see \collisionrate{} in Table~\ref{tab:analysis:adv-testing}) suggests that there exist some patterns among these adversarial inputs. One possible pattern that we set to analyze is a spatial correlation among adversarial inputs in different models. Differently put, we were interested in understanding whether \emph{adversarial inputs concentrate in similar areas in the agent's input space.}

\begin{figure}[t]
\centering
\includegraphics[width=1\linewidth]{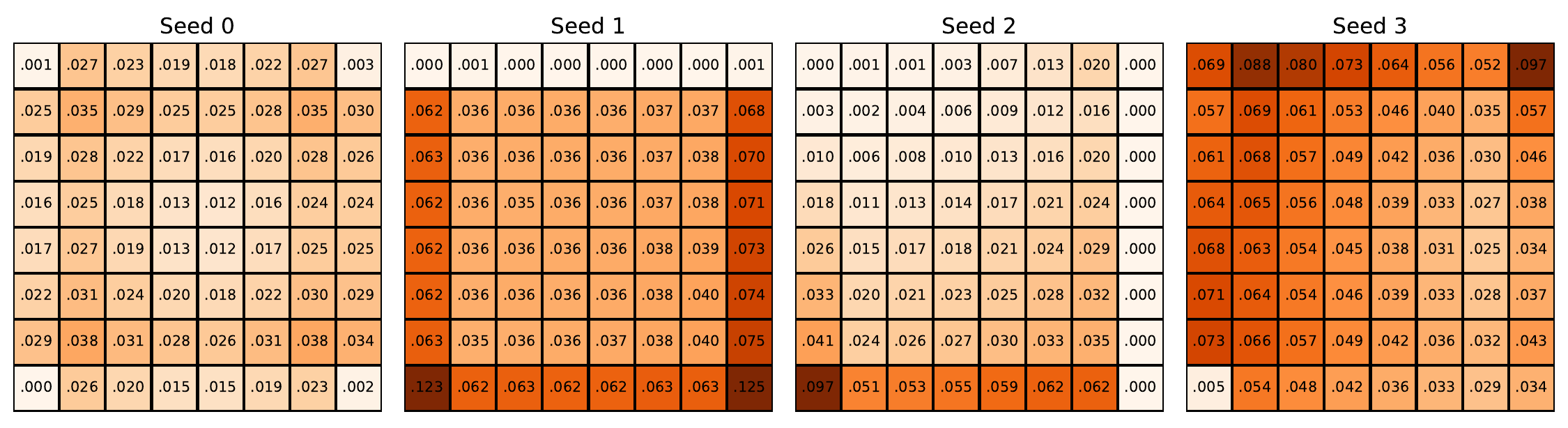}
\caption{\gridenv{}: 
A heatmap that highlights the concentration of adversarial inputs (identified via the  \adversarialrate{} metric) per each cell.}
\label{fig:analysis:spatial}
\end{figure}

\medskip
\noindent
\textbf{Evaluation.}
Once again, the simplified \gridenv{} environment is extremely beneficial for such an analysis, as this environment includes states that are (i) symmetric, and (ii) include randomly assigned obstacles that are uniformly sampled. These properties assist in eliminating many unwanted biases that can arise in other DRL environments, hence skewing any such spatial analysis. In other words, all the cells of the grid are equally challenging for the agent. 
We chose the $5$ best seeds in terms of success rate (i.e., more than $95\%$ in our empirical evaluation) and validated that these gave rise to agents that did not collide in any of the training episodes. All models were trained with the same configuration and hyperparameters, differing only in the random initialization of the parameters.
Next, we divided the (continuous) input space into subregions and used \prove{} to assess the ratio of adversarial inputs (and effectively, the \adversarialrate{}) in each subregion. Surprisingly, although all models that passed our performance threshold seemed identical (as well as safe) during the thorough evaluation phase, their susceptibility to adversarial inputs differed \emph{significantly}. In addition, we also observed high variability in the \emph{spatial localization} of regions in the input space that includes adversarial inputs. These results are highlighted in Fig.\ref{fig:analysis:spatial}.

\medskip
\noindent
\textbf{Implications. } 
Our evaluation leads to an important conclusion regarding the spatial properties of adversarial inputs --- separate models trained in the same environment, with the same algorithm, hyperparameters, and even with an identical success rate, may have vast differences in their susceptibility to adversarial inputs. In practice, this can be translated to high variability in the heat maps representing their \adversarialrate{} across our evaluation. This also suggests that we cannot directly deduce the concentration of adversarial inputs to one model, based on its peers. Furthermore, these results indicate that the adversarial inputs seem to be an inherent property of the neural network function rather than being influenced by environmental factors. While the finding that adversarial inputs are independent of the dataset is not novel in the literature \citep{szegedy2013intriguing}, as far as we are aware, our work presents the first empirical demonstration of this phenomenon in the context of deep reinforcement learning.

\subsection{Temporal Analysis of Adversarial Inputs}
The previous analysis indicated that indeed, different models with nearly identical characteristics (training and environment) may have unsafe regions in which the agent may demonstrate unwanted behavior. However, another interesting question is whether, for a given model, these unsafe regions are constant throughout the training phase. In other words, for a fixed model, \emph{do the unsafe regions change during training}? This analysis is important not only for understanding the effect that the training phase has on the model's vulnerability to adversarial inputs, but also, to delve into the question of how various training algorithms can affect the model's susceptibility.

\begin{figure}[ht]
\centering
\includegraphics[width=1\linewidth]{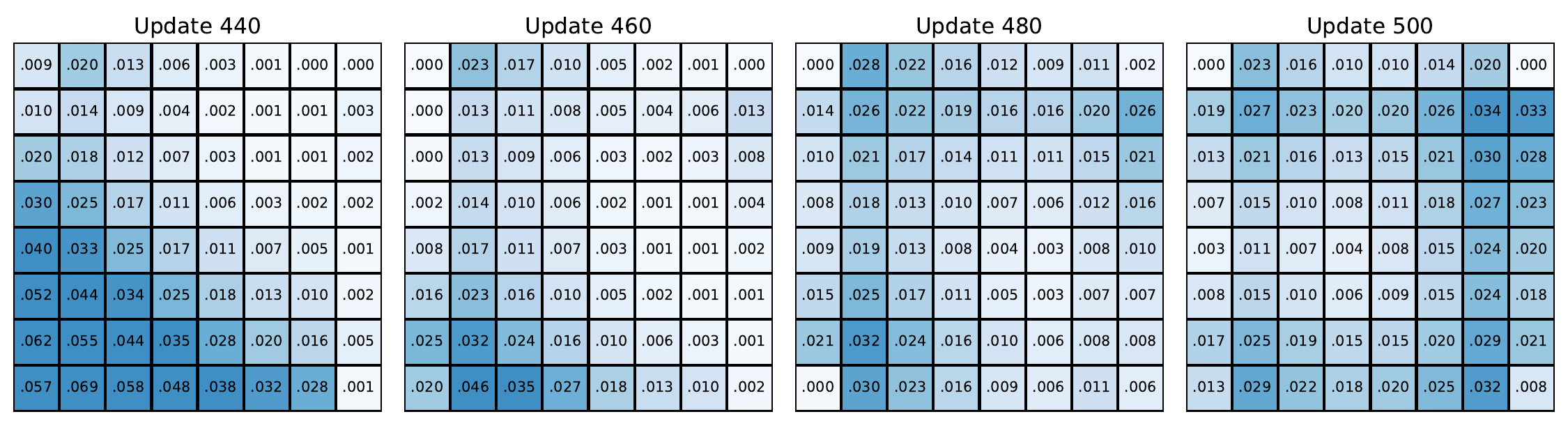}
\caption{\gridenv{}: A qualitative analysis of the temporal distribution.}
\label{fig:analysis:temporal-a}
\end{figure}

\medskip
\noindent
\textbf{Evaluation.}
We generated a fresh set of $5$ models, each based on a unique random seed value, used to generate the initial weights and biases. We then conducted a full training process over $500$ episodes, finally selecting the best-performing model; after this process, we analyzed $5$ snapshots from the last $60$ training steps with an interval of $20$ episodes (recall that formal methods are crucial for this part, as all models performed quite similar empirically and it is hard to distinguish them one from the other). Our results, summarized in Fig.~\ref{fig:analysis:temporal-a}, demonstrate that the position of unsafe regions can drastically change during the training process, confirming the results obtained in the previous analysis. 

\medskip
\noindent
\textbf{Implications.} 
Our findings lead to an important conclusion: it is very challenging to solve the susceptibility of an agent to adversarial inputs, during training. Specifically, a main challenge is that occurrences of unsafe regions ``shift"  during training; hence, even if we were to identify such an unsafe region at a given time-step, and potentially solve the problem locally, this process can generate additional unsafe regions in unexplored areas. In a worst-case scenario, this shifting of unsafe regions can continue ad infinitum. Importantly, however, we note that this does not necessarily imply that safe training is an unsolvable problem, but rather, that it is a challenging task that opens up several future research directions, which are currently pursued (as we elaborate further in Sec.~\ref{sec:conclusion}).

The first implication is an additional confirmation for the conclusion of our spatial analysis. Namely, our results indicate that even \emph{slightly} different models (differing only in a handful of training epochs) may have completely different unsafe regions, relative to the model in question.
The second implication, as demonstrated by our results (summarized in Fig.~\ref{fig:analysis:temporal-a}) is that even if the training phase converges to a seemingly successful policy (usually depicted by a plateau in the agent's learning curve), there may be a high variability among the unsafe regions and thus in the safety of these agents. Hence, in the context of safety-critical systems, it is crucial to evaluate the model not only empirically, but also to exploit additional metrics, as governed by formal verification techniques.

\subsection{Relating Adversarial Examples to the DNN Architecture}

In the previous subsections, we focused on the spatial and temporal concentration of adversarial examples in the input space, specifically demonstrating their abundance even in state-of-the-art DRL training processes. Next, we focus on an additional, complementary aspect of adversarial examples. Namely, we focus on the question of \textit{if, and how, the DNN architecture correlates to adversarial examples}? This analysis is important, as it can shed light on architectures that should be avoided, in order to mitigate the chance of posing the agent to be susceptible to such attacks. Specifically, we focus on two main factors, the (i) size, and (ii) activation functions, characterizing a DNN in question. We also raise these issues to justify, or refute, the common folklore belief, that very expressive DNNs (i.e., DNNs with many parameters) are more susceptible to adversarial inputs, compared to their less expressive peers.

\begin{table}[t]
\centering
\begin{tabular}{|l|c|c|c|c|c|}
\hline
 & 2x32 & 4x128 & 4x256 & 6x256 \\
\hline
\texttt{Jumping W.} & $0.09\pm0.04$ & $0.12\pm0.01$ & $0.17\pm0.03$ & $0.20\pm0.03$\\
\texttt{Navigation} & $1.69\pm1.35$ & $2.14\pm1.03$ & $2.52\pm1.58$ & $4.59\pm0.73$ \\
\hline
\end{tabular}
\caption{Average \adversarialrate{} of the $4$ best-performing models obtained with the corresponding network sizes on the two considered environments.}
\label{tab:analysis:adv-testing-size}
\end{table}

\begin{figure}[b]
\centering
\includegraphics[width=1\linewidth]{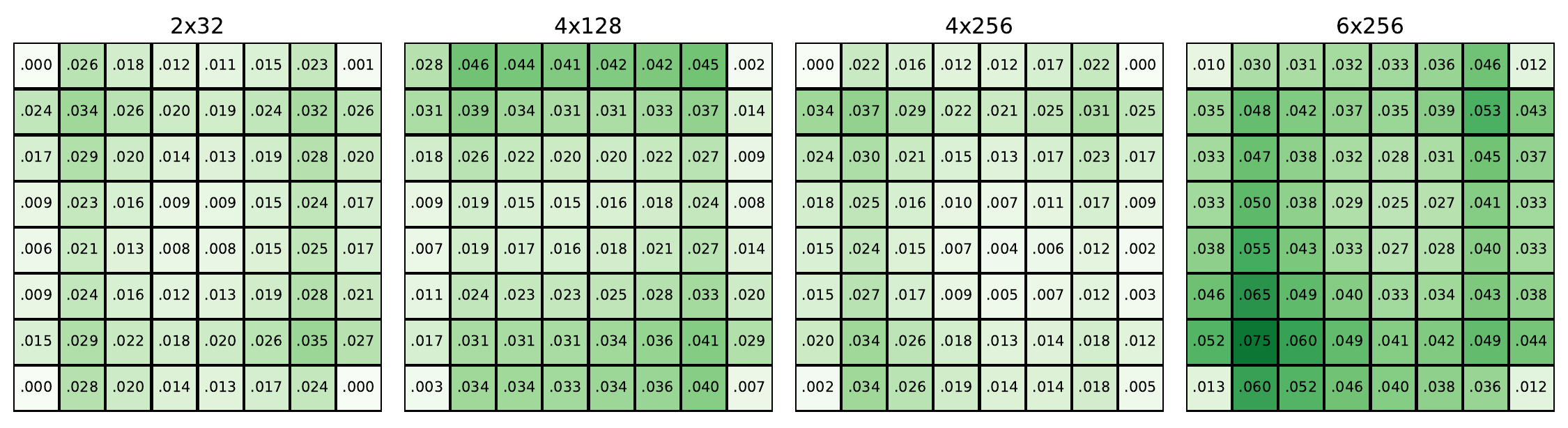}
\caption{\gridenv{}: The unsafe regions computed on the best model, obtained with $4$ varying sizes of the neural network. }
\label{fig:analysis:size}
\end{figure}

\medskip
\noindent
\textbf{Evaluation: DNN Size.}
In our first experiment, we trained models with various sizes, ranging from $32$ to $528$ hidden neurons in total, as well as $2$ to $6$ hidden layers. As in the previous experiments, we retained only the models that performed successfully, with an average reward that is above our $95\%$ threshold, and thus demonstrated similar performance in terms of \successrate{} and \collisionrate{}. We repeated this experiment for both the \gridenv{} and \turtleenv{} environments.
Then, we calculated the obtained average \adversarialrate{} for all architectures, across both benchmarks, as reported in Table~\ref{tab:analysis:adv-testing-size}. As we hypothesized, the results indicate a clear correlation between the DNN's size and its susceptibility to adversarial inputs. The larger the DNN, the more it tended to be susceptible to adversarial regions, and the more concentrated these appeared to be in the input space.
Fig.~\ref{fig:analysis:size} depicts these results for $4$ varying network sizes, w.r.t. agents trained on the \gridenv{} benchmark.

\begin{figure}[t]
\centering
\includegraphics[width=1\linewidth]{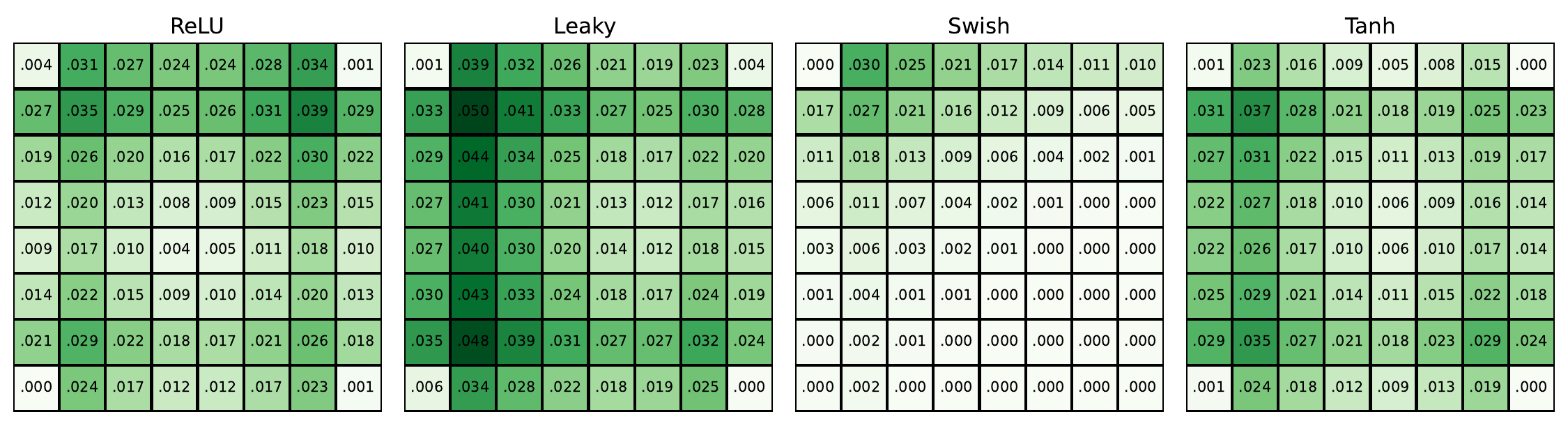}
\caption{\gridenv{}: The unsafe regions computed on the best model obtained, with $4$ varying activation function types.}
\label{fig:analysis:activation}
\end{figure}

\medskip
\noindent
\textbf{Evaluation: DNN Activation Functions.}
Another aspect of interest is the correlation of the activation function \emph{type}, on the existence of adversarial inputs. Broadly speaking, we set out to understand whether two DNNs of the same size (i.e., identical depth and number of neurons) can demonstrate varying sensitivity levels to adversarial inputs, solely due to having different activation functions. To this end, we constructed the same aforementioned experimental setup, with the difference of selecting the models based on $4$ types of activation functions as described in Table~\ref{tab:analysis:adv-testing-activation}. Interestingly, we found no clear correlation between these two factors, as depicted in Fig.~\ref{fig:analysis:activation}, which presents our results on the \gridenv{} benchmark \footnote{Notice that models with the \textit{sigmoid} activation function failing to reach the cutoff of $95\%$ success in our experiments. Still, we reported these results for completeness but excluded them in our further analysis.}, confirming the results of Table~\ref{tab:analysis:adv-testing-activation}. We also note that, although our results showcase that the \texttt{Swish} activation function provides better results on the \gridenv{} environment, these results are not replicated on the (more complex) \turtleenv{} benchmark. In the latter case study, there is no evidence of any correlation between the type of activation function and the size and abundance of the unsafe regions.

\medskip
\noindent
\textbf{Implications.}  
Our evaluation indicates that specific characteristics of a DNN architecture may increase the abundance of adversarial inputs, while other features do not have an apparent effect.
Specifically, we observed a strong correlation between the \emph{size} of the network and the presence of adversarial inputs and unsafe regions. This implies that for safety-critical tasks in DRL, small neural networks may be preferable to larger ones. On the other hand, there does not seem to be a clear correlation between the \emph{type} of the activation functions and the \adversarialrate{} value. This implies that when conducting the DRL training --- it is preferable, in general, to greedily select an activation function that is associated with high rewards, on average, without the need to take the adversarial rate into account. 

\begin{table}[ht]
\centering
\begin{tabular}{|l|c|c|c|c|c|}
\hline
 &\texttt{ReLU} & \texttt{Tanh} & \texttt{Sigmoid} & \texttt{Swish} & \texttt{Leaky}  \\
\hline
\texttt{Jumping W.} & $0.05\pm0.02$ & $0.11\pm0.04$ & - & $0.02\pm0.02$ & $0.03\pm0.01$ \\
\texttt{Navigation} & $1.85\pm0.54$ & $1.05\pm0.98$ & - & $2.84\pm0.13$ & $1.66\pm0.24$ \\
\hline
\end{tabular}
\caption{Average \adversarialrate{} of the $5$ best-performing seeds obtained with the corresponding activation function on the two considered environments.}
\label{tab:analysis:adv-testing-activation}
\end{table}

\subsection{Extension to the Mapless Navigation Case-Study}
In the first analysis of this section, we performed our evaluation on the \gridenv{} environment while in this section, as a proof of concept, we replicate the experiments on the more complex and realistic \turtleenv{} task. This analysis is not as straightforward as the previous one for multiple reasons: (i) in a real-world scenario, it is hard to encode the complete set of requirements that guarantees a zero-violation behavior, however, we believe this requirement goes beyond the scope of this work and we thus focus on specific safety properties that are informative enough for our purposes; (ii) the \turtleenvshort{} is a well-known and challenging problem in the literature \citep{tai2017virtual}, and obtaining a model that does not collide in an extensive empirical evaluation is a complex task that requires delicate techniques and fine-tuning hyperparameters \citep{marchesini2022enhancing}. Once again, this goes beyond the objective of this work, which does not focus on the training aspects but rather on the offline evaluation; for this reason, we impose a cutoff of $3\%$ on the \collisionrate{} and $94\%$ on the \successrate{}. 

Tab. \ref{tab:analysis:adv-testing-navigation} depicts our results. Even considering the previously described limitations of the environment, it is clear that the pattern highlighted in our running example is confirmed also in this real-world robotic problem. In particular, the empirical \collisionrate{}, which is relatively low, drastically increases when the evaluation is performed in the proximity of the adversarial inputs (i.e., \fvcollisionrate{}); crucially, the \adversarialrate{} is always greater than zero, even in our best-performing model. In general, the results of Tab.~\ref{tab:analysis:adv-testing-navigation} confirm our previous findings, corroborating the hypothesis that vulnerability to adversarial inputs is not domain-specific.

\begin{table}[t]
\centering
\begin{tabular}{|c|c|c|c||c|}
\hline
Seed & \successrate & \collisionrate & \adversarialrate & \fvcollisionrate \\
\hline
0 & $95\pm0.7$ & $2\pm1.1$ & $1.4$ & $23\pm4.3$ \\
1 & $94\pm1.2$ & $1\pm0.7$ & $1.9$ & $17\pm2.5$ \\
2 & $94\pm0.7$ & $2\pm1.3$ & $2.8$ & $21\pm3.4$ \\
3 & $98\pm0.4$ & $1\pm0.4$ & $2.1$ & $20\pm3.5$ \\
4 & $95\pm1.1$ & $1\pm0.9$ & $0.8$ & $19\pm2.4$ \\
\hline
\end{tabular}
\caption{Comparative analysis between simple empirical testing and adversarial testing, performed in the proximity of the violating points identified via \prove.}
\label{tab:analysis:adv-testing-navigation}
\end{table}

\begin{table}[b]
\centering
\begin{tabular}{|c|c|c|c||c||c|}
\hline
Seed & \texttt{Adv.Collision Rate} & \texttt{Cross-Seed Adv.Collision Rate} \\
\hline
1 & $17\pm2.5$ & $2\pm0.7$ \\
2 & $21\pm3.4$ & $2\pm1.3$ \\
3 & $20\pm3.5$ & $3\pm0.2$ \\
4 & $19\pm2.4$ & $1\pm0.7$ \\
\hline
\end{tabular}
\caption{The \fvcollisionrate{} is computed by sampling the initial states in the proximity of the adversarial inputs of the model generated by \textit{seed 0}, with regard to the other models.}
\label{tab:analysis:adv-testing-navigation-ablation}
\end{table}

\medskip
\noindent
\textbf{Ablation Study: Cross-Seed Adversarial Collision Rate.}  
We also performed an ablation study, with the aim of providing additional proof to the presented results. Specifically, we performed an empirical evaluation in the proximity of the adversarial inputs found for the first  model (i.e., seed $0$) with regard to all the other models. Our results (summarized in Tab. \ref{tab:analysis:adv-testing-navigation-ablation}) show that in this case, the \collisionrate{} does not increase. We believe this is additional evidence to the claim that adversarial inputs are a specific property of each model independently. Hence, it is not possible to rely on the formal analysis of a single model, w.r.t. its adversarial inputs, in order to infer the safety of other models. Note that we performed this analysis on the \turtleenv{} benchmark due to the relatively high average collision rate that characterizes the models trained on this task which, in turn, emphasizes the differences among the models.

\subsection{Training with TD3}
Finally, we replicated the previous experiments on a fresh batch of models trained with a different algorithm --- TD3, instead of PPO \citep{fujimoto2018addressing}. The objective of this additional evaluation is to demonstrate that the problems raised are not unique to PPO, but are rather shared among different algorithms and training approaches. Towards this end, we rely on the DDPG family of training methods \citep{schulman2015high}. Unlike PPO, which is a policy gradient approach, TD3 represents a value-based alternative, and thus an interesting variant of our experiments to explore. Another advantage of using TD3, instead of the popular DQN \citep{mnih2013playing} or Rainbow \citep{hessel2018rainbow}, is that it is not limited to discrete action spaces. The choice of the models and the setting for the empirical evaluation follows the setup of the previous analysis. Tab.~\ref{tab:analysis:adv-testing-ddpg} presents our results, that confirm the findings of the previous analysis on PPO. We believe that, once again, this result corroborates the hypothesis that adversarial inputs are agnostic to the DRL training algorithm used in practice. 

\begin{table}[h]
\centering
\begin{tabular}{|c|c|c|c||c|}
\hline
Seed & \successrate & \collisionrate & \adversarialrate & \fvcollisionrate \\
\hline
0 & $98\pm1.0$ & $2\pm1.5$ & $6.5$ & $34\pm2.4$ \\ 
1 & $96\pm1.7$ & $1\pm0.4$ & $1.0$ & $12\pm1.3$ \\ 
2 & $95\pm1.5$ & $1\pm0.3$ & $2.1$ & $19\pm2.1$ \\ 
3 & $98\pm0.7$ & $1\pm0.6$ & $4.5$ & $26\pm2.2$ \\ 
\hline
\end{tabular}
\caption{Extension of the experiments of Tab. \ref{tab:analysis:adv-testing-navigation} for the models trained with the TD3 algorithm.}
\label{tab:analysis:adv-testing-ddpg}
\end{table}
\section{Training and Retraining Approaches}
\label{sec:retraining}

The subject matter of this paper is leveraging verification-based techniques to better understand the impact of adversarial configurations and their behavior. However, we note that a complementary research direction consists of directly generating models that are robust against such perturbations by exploiting specifically designed training techniques. For example, in the work of \cite{yang2022neural}, the authors propose to iteratively identify the unsafe regions, for a given set of safety requirements, and exploit a retraining process to repair them. In particular, the authors propose to initialize the new training episodes in proximity to various adversarial inputs found during the offline analysis. Although presenting significant results, this work, like most training-driven techniques, does not provide formal guarantees. In addition, our results show that the adversarial inputs are prone to ``shift'' during training, and thus focusing on a particular region of the domain may only result in moving the unsafe region to another part of the input domain.
Alternative approaches focus on \emph{directly} repairing the weights of the neural network to manually patch the adversarial points. Although this method, and similar techniques, are effective in some contexts, they are mostly geared towards classification models, and thus inadequate to many DRL agents that typically learn probability distributions over the possible actions. An alternative promising future direction involves the employment of constrained techniques, that restrict the search space of the policies inside safe bounds. This family of algorithms includes popular techniques, such as CPO \citep{achiam2017constrained}, SOS \citep{marchesini2022exploring}, Lagrangian-PPO \citep{ray2019benchmarking}, and IPO \citep{liu2020ipo}. These methods can complement our approach by integrating with the computation procedure of the \adversarialrate{}, and by adding an additional objective during training, aimed at minimizing it. However, constrained DRL algorithms typically provide results that are guaranteed only in expectation, hence failing to consider various small input configurations that are tailored by a potential adversary.

\section{Conclusion}
\label{sec:conclusion}

In this paper, we focus on the timely problem of safety and reliability among DRL agents. Specifically, we delved into the enigmatic nature of adversarial inputs, which are a well-known vulnerability among DNNs and limit their deployment in real-world systems. First, we presented novel tools and algorithms to formally and rigorously evaluate the vulnerability of state-of-the-art DRL agents. In particular, \textcolor{black}{we build upon our previous work, and in particular on \texttt{ProVe}~\citep{ProVe}, extending the \texttt{Violation Rate} metric introduced therein into the \adversarialrate{}: a metric specifically adapted for the systematic evaluation of adversarial inputs in DRL. The primary contribution of this work is therefore a comprehensive evaluation framework for studying the effect of adversarial inputs on DRL policies, rather than a new verification tool per se.} The \adversarialrate{} allows us to quantify the susceptibility of DRL agents to various input perturbations, and learn about their characterization in time and space.
In addition, we also demonstrated how this metric can be used to characterize various aspects of adversarial inputs, that have yet to be sufficiently covered by the literature. Moreover, we provide a series of extensive experiments, that suggest that for various models, benchmarks, and training algorithms --- the susceptibility to adversarial inputs is an inherent problem. We also demonstrated how such adversarial inputs are dispersed in the input space, and how the architecture of a given model can affect its robustness, or lack thereof. We believe that this work can have a significant impact on the DRL community, especially with regard to applying these techniques in real-world systems, in which we must ensure robustness. 
We hope this work will pave the way for additional research on DRL safety, with both theoretical and practical aspects w.r.t. analysis of safety risks.
\textcolor{black}{We also note that the spatial characterization provided by the \adversarialrate{} opens an interesting direction for the design of \textit{localized safety shields}~\citep{marchesini2022exploring}: a contiguous unsafe region can be protected by a shield that intervenes only in that specific subregion, minimizing the impact on overall performance, whereas scattered adversarial inputs would require a more pervasive strategy.}
%


\medskip
\noindent
\textbf{Acknowledgements.} 
The work of Corsi and Amir was carried out during their time at the University of Verona and the Hebrew University of Jerusalem, respectively.
The work of Corsi and Farinelli was carried out within the Interconnected Nord-Est Innovation Ecosystem (iNEST) and received funding from the European Union Next-GenerationEU (PIANO NAZIONALE DI RIPRESA E RESILIENZA (PNRR) – MISSIONE 4 COMPONENTE 2, INVESTIMENTO 1.5 – D.D. 1058 23/06/2022, ECS00000043). The work of Amir was supported by a scholarship from the Clore Israel Foundation during his doctorate. The work of Amir and Katz was partially funded by the European Union (ERC, VeriDeL, 101112713). Views and opinions expressed are however those of the author(s) only and do not necessarily reflect those of the European Union or the European Research Council Executive Agency. Neither the European Union nor the granting authority can be
held responsible for them.

{
\bibliographystyle{abbrv}
\bibliography{bibliography}
}

\end{document}